\documentclass[lettersize,journal]{IEEEtran}
\usepackage{amsmath,amsfonts}
\usepackage{algorithmic}
\usepackage{algorithm}
\usepackage{array}
\usepackage[caption=false,font=normalsize,labelfont=sf,textfont=sf]{subfig}
\usepackage{textcomp}
\usepackage{stfloats}
\usepackage{url}
\usepackage{verbatim}
\usepackage{graphicx}
\usepackage{cite}
\usepackage{multirow}
\usepackage{booktabs}

\begin{document}

\title{Not Only Consistency: Enhance Test-Time Adaptation with Spatio-temporal Inconsistency for Remote Physiological Measurement}

\author{Xiao Yang, Jiyao Wang~\IEEEmembership{Student Member,~IEEE}, Yuxuan Fan, Can Liu, Houcheng Su, Weichen Guo, Zitong Yu~\IEEEmembership{Senior Member,~IEEE}, Dengbo He, Kaishun Wu~\IEEEmembership{Fellow,~IEEE}
        
\thanks{This manuscript was first submitted in August. 22 2025. (Corresponding authors: Jiyao Wang, Dengbo He).}

\thanks{Xiao Yang, Jiyao Wang and Dengbo He are with the Systems Hub, the Hong Kong University of Science and Technology (Guangzhou), Guangzhou, China,
(e-mail: xyang856@connect.hkust-gz.edu.cn; jwanggo@connect.ust.hk; dengbohe@hkust-gz.edu.cn).}
\thanks{Yuxuan Fan, Houcheng Su and Kaishun Wu are with the Information Hub, the Hong Kong University of Science and Technology (Guangzhou), Guangzhou, China, (e-mail: yfan546@connect.hkust-gz.edu.cn; hsu638@connect.hkust-gz.edu.cn; wuks@hkust-gz.edu.cn).}
\thanks{Can Liu and Weichen Guo are with the School of Information Engineering, Sichuan Agricultural University, Yaan, China, (e-mail: 202203742@stu.sicau.edu.cn; guoweichen@stu.sicau.edu.cn).}
\thanks{Zitong Yu is with the School of Computing and Information Technology,
Great Bay University, Dongguan 523000, China, also with the National Engineering Laboratory for Big Data System Computing Technology, Shenzhen
University, Shenzhen 518060, China, and also with Dongguan Key Laboratory
for Intelligence and Information Technology, Dongguan 523000, China (e-
mail: yuzitong@gbu.edu.cn).}}

\markboth{Journal of \LaTeX\ Class Files,~Vol.~14, No.~8, August~2021}%
{Shell \MakeLowercase{\textit{et al.}}: A Sample Article Using IEEEtran.cls for IEEE Journals}


\maketitle

\begin{abstract}
Remote physiological measurement (RPM) has emerged as a promising non-invasive method for monitoring physiological signals using the non-contact device. Although various domain adaptation and generalization methods were proposed to promote the adaptability of deep-based RPM models in unseen deployment environments, considerations in aspects such as privacy concerns and real-time adaptation restrict their application in real-world deployment. Thus, we aim to propose a novel fully Test-Time Adaptation (TTA) strategy tailored for RPM tasks in this work. Specifically, based on prior knowledge in physiology and our observations, we noticed not only there is spatio-temporal consistency in the frequency domain of BVP signals, but also that inconsistency in the time domain was significant. Given this, by leveraging both consistency and inconsistency priors, we introduce an innovative expert knowledge-based self-supervised \textbf{C}onsistency-\textbf{i}n\textbf{C}onsistency-\textbf{i}ntegration (\textbf{CiCi}) framework to enhances model adaptation during inference. Besides, our approach further incorporates a gradient dynamic control mechanism to mitigate potential conflicts between priors, ensuring stable adaptation across instances. Through extensive experiments on five diverse datasets under the TTA protocol, our method consistently outperforms existing techniques, presenting state-of-the-art performance in real-time self-supervised adaptation without accessing source data. The code will be released later.
\end{abstract}


\section{Introduction}
\IEEEPARstart{W}{ith} the advancement of remote physiological measurement (RPM) technology, physiological data such as heart rate (HR), heart rate variability (HRV), and blood volume pulse (BVP) can now be estimated using various non-contact multimedia devices \cite{liu2024rppg, zhang2024self,magdalena2018sparseppg, choi2024fusion}, for example, RGB cameras and millimeter-wave (mmWave) radar. This technology holds potential for applications across various fields, particularly in human-computer interaction \cite{wang2024multi, wang2024revisiting}, affective computing \cite{lu2024gpt}, and state monitoring \cite{wang2024efficient, gong2024heart}. Particularly, deep learning-based RPM methods have gained attention in recent years \cite{petrovic2019high,niu2020video,wang2024efficient}. However, they struggle with generalization due to variations existing in different datasets (e.g., devices latency differences \cite{sun2023resolve}, illumination and recording environment \cite{wang2023hierarchical}). To address this issue, the domain generalization (DG) \cite{lu2023neuron,wang2023hierarchical} and domain adaptation (DA) \cite{du2023dual} were explored. However, unavailable target domain labels make DG methods only converge at a domain-agnostic optimal point, they lack target domain specificity \cite{wang2023hierarchical}. Although DA methods allow access and adaptation to target domain data, we cannot obtain all the data of the target domain at once in the real world \cite{hsieh2022augmentation}. Therefore, fully Test-Time Adaptation (TTA) has emerged recently \cite{huang2024fully,li2024bi}, enabling real-time model to realize self-supervised fine-tuning on target domains during inference without needing offline source data. These approaches address the need for real-time performance improvement in unseen domains while simultaneously satisfying privacy requirements by eliminating the need for direct access to source domain data and target domain labels.


\begin{figure*}[h] 
    \centering
    \includegraphics[width=\textwidth]{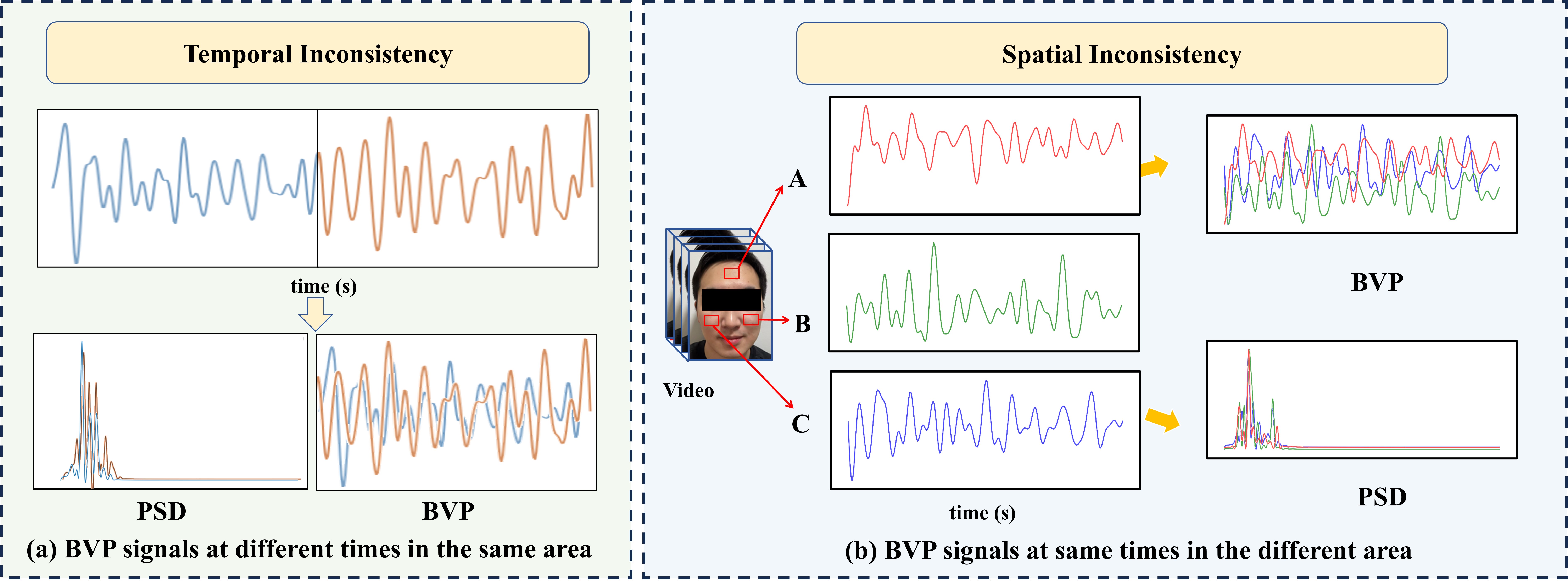} 
    \caption{Illustration of Spatio-temportal Inconsistency in BVP signals. (a) shows a continuous BVP signal from the same facial region, where the power spectral density (PSD) of the two consecutive segments is similar, yet there are differences in signal morphology. (b) demonstrates that within the same timeframe, different facial regions exhibit similar PSDs, but the signal morphology also varies.}
    \label{fig:ts_inconsistency}
\end{figure*}

Nevertheless, previous TTA and self-supervised methods for RPM primarily focus on one prior knowledge, called spatio-temporal consistency, to facilitate model convergence \cite{sun2022contrast}. It elaborates that BVP signals reflected within a short time interval exhibit similar power spectral density (PSD), as well as the similarity of PSD across different facial regions at the same time. However, based on our observation (as illustrated in Fig. \ref{fig:ts_inconsistency}), except for the consistency in the frequency domain, there are morphological differences in the time domain of BVP signals sampled from different facial regions and time slots. Specifically, since cardiac activity and physiological processes are inherently complex, the amplitude and offset of heartbeats change at every moment \cite{schandry1981heart}. Additionally, due to the absorption of blood oxygen by body tissues and the varying distances of these regions from the heart, the BVP signal extracted from different facial regions also presents differences in pulse transit time and waveform \cite{graaff1993optical,smith1999pulse}. Unfortunately, in past research, such inherent variability in BVP was ignored or eliminated by the widely used spatio-temporal consistency. 

Inspired by this, we introduce a novel \textbf{C}onsistency-\textbf{i}n\textbf{C}onsistency-\textbf{i}ntegration (\textbf{CiCi}) framework utilizing both spatio-temporal frequency-domain consistency (\textbf{STFC}) and time-domain inconsistency (\textbf{STTI}) characteristics of BVP signals to offer effective supervision in the TTA process. Specifically, based on an analysis of cardiac activity and observations of BVP signals, we first leverage the STFC prior to regularize the PSD of the BVP signal corresponding to a specific HR section that remains stable with the spatio-temporal changes. Besides, a novel self-supervised loss based on STTI prior is supplemented to provide effective alternative supervision. The STTI prior ensures the time-domain morphological difference is not neglected while keeping the frequency-domain semantic consistency. Furthermore, since there is no ground truth for supervision and the optimization direction of two priors is sort of in conflict, the convergence of the model will be difficult as it can result in significant deterioration in model performance and disordered output signals. To address this issue, we design a gradient dynamic control (\textbf{GDC}) strategy to mitigate the conflict gradient propagation directions of STFC and STTI. Adaptive weights are distributed to two regularizations depending on the relative magnitude of their gradients when an orthogonality between the gradients of the two priors is detected. It restricts the conflict at a proper level while maintaining both the consistency and inconsistency behind the BVP signal. In all, our contributions are summarized as follows:

\begin{itemize}
    \item We leverage and instantiate a novel prior knowledge based on the inherent morphological inconsistencies in the time domain.
    \item We designed an effective TTA framework Consistency-inConsistency-integration (CiCi) combining spatio-temporal frequency-domain consistency and time-domain inconsistency priors to provide appropriate supervision from different perspectives.
    \item We propose a gradient dynamic control strategy to eliminate the model degradation caused by the conflict optimization directions of two priors, providing a stable and effective adaptation direction.
    \item Through extensive experiments across five RGB datasets and two mmWave datasets under the fully TTA protocol, our method achieved superior performance compared to previous methods, validating the effectiveness of our approaches.
\end{itemize}

\section{Related Work}
\subsection{Remote Physiological Measurement}
Early research primarily focused on exploring the feasibility of extracting heart rate from skin color fluctuations using standard cameras \cite{verkruysse2008remote,poh2010non,de2013robust}. This camera-based physiological monitoring method is known as remote photoplethysmography (rPPG). Additionally, the frequency changes in the mmWave signals reflected from chest movements, used to estimate physiological information \cite{gu2013doppler,wang20131}, have also been widely applied in RPM field. In recent years, with remarkable advancements in computer vision and computational capabilities, deep learning methods have increasingly been applied in remote physiological estimation \cite{niu2019rhythmnet,liu2020multi,wang2024condiff, choi2024fusion}. These approaches have demonstrated unprecedented accuracy in extracting not only HR but also other vital signs such as breathing rate, HRV, and blood oxygen saturation \cite{wang2025physmle}.
 
Despite their excellent modeling capabilities and promising results in controlled settings, these DL-based RPM models often have limited generalization capacity when deployed in real-world environments \cite{cheng2021deep}. Critical challenges include varying illumination conditions, subject motion states, diverse skin tones, and external environmental factors, compounded by inherent delays in acquisition devices and the predominant use of lab-based training data \cite{lee2024review}. To address these cross-domain challenges, recent RPM estimation methods have made significant strides by adopting advanced DG \cite{lu2023neuron,sun2023resolve,wang2023hierarchical,wang2024tim} and DA \cite{du2023dual} strategies. These approaches aim to bridge the gap between controlled environments and real-world applications through advanced transfer learning and feature extraction. While DL-based RPM methods have made progress, DG lacks target specificity \cite{wang2025align}, and DA needs impractical labeled data from both the source and target domains \cite{xie2024sfda}. To address these issues, we focus on fully TTA, allowing real-time model to self-supervise fine-tuning during inference without offline source data.

 \subsection{Test-Time Adaptation}

TTA aims to adapt a pre-trained model to the target domain at the testing phase using only input test samples from the target domain \cite{zhao2023delta,wang2020tent}. Current TTA methods can primarily be grouped into three categories: Batch Normalization (BN) calibration \cite{schneider2020improving,nado2020evaluating,duboudin2022learning,bahmani2022semantic,klingner2022continual}, meta-learning approaches \cite{min2023meta,liu2022towards,chi2021test,soh2020meta}, and self-supervised learning \cite{zhang2020inference,li2021test}. BN calibration methods adjust batch normalization statistics to better reflect the target domain, while meta-learning approaches \cite{min2023meta,liu2022towards,chi2021test,soh2020meta} enable rapid adaptation to new tasks with limited samples and gradient updates. On the other hand, self-supervised learning methods, such as model fine-tuning \cite{zhang2020inference,li2021test}, leverage test mini-batches for adaptation. However, these TTA methods generally require aggregating a large number of samples from the target data to form a batch. For continuous regression tasks like RPM, this often necessitates a redesign of the approach.

More recently, TTA began to gain attention in RPM tasks. Huang et al. \cite{huang2024fully} proposed a feature learning method guided by synthetic signals to enhance the adaptability of rPPG models to new target domains. However, this approach incurs additional memory overhead. Li et al. \cite{li2024bi} were the first to apply TTA methods to the rPPG task, requiring no modifications to network architecture and leveraging self-supervised consistency priors to improve model performance in unseen domains. Although consistency priors have been widely utilized in many self-supervised rPPG methods \cite{xie2024sfda, wang2025physmle}, the objectives of TTA differ from those of DG. DG methods focus on identifying invariants across domains to boost generalization \cite{wang2023hierarchical}, whereas TTA focuses more on adaptation to specific scenes. If TTA only emphasizes consistent information, it may overlook unique scene details. Thus, our work highlights the importance of the inherent variability of the BVP signal for effective self-supervised adaptation tailored to each scene. 

\section{Observations about BVP Signals}
\label{section: Observations About rPPG}

\subsection{Temporal Inconsistency}

As shown in Fig. \ref{fig:ts_inconsistency} (a), aligned with previous studies \cite{sun2022contrast}, we noticed that the power spectral density (PSD) of two segments of one continuous BVP signal with a short time interval is similar (i.e., the close HR value). This indicates they exhibit similar HR values in the frequency domain. However, from the perspective of the time domain, their shapes and amplifications are not consistent. Multiple physiological factors, such as respiration \cite{persson1996modulation}, vascular tension in large and microvessels \cite{johnson1986nonthermoregulatory, heistad1973interaction}, and thermoregulation \cite{ivanov2014network}, cause oscillations in blood volume within tissues. 

Despite we can use the ground-truth BVP signal to make the model learn the appropriate shape (e.g., using Pearson coefficiency as loss function \cite{wang2023hierarchical, lu2023neuron}) in supervised learning, the lack of proper morphology constraints and only regularizing the temporal consistent in the frequency domain will result in the improper output preference of the model in self-supervision (i.e., the model will prefer to generate similar BVP signal shape when there is the same HR value).

\subsection{Spatial Inconsistency}
Even in ideal conditions, ignoring the delays caused by circuit design differences between acquisition devices, BVP signals from different facial regions exhibit morphological variability, primarily due to five reasons: 1) The optical properties of skin vary across different regions, resulting in differing rates of light absorption and reflection \cite{graaff1993optical}; 2) differences in blood flow and vascular distribution across facial areas lead to varying degrees of blood volume changes affecting light reflection \cite{lister2012optical}; 3) the varying distances of facial regions from the heart lead to differences in pulse transit time \cite{smith1999pulse}, causing asynchrony in rPPG signals across regions \cite{sun2023resolve}.

As shown in Fig. \ref{fig:ts_inconsistency}(b), although the PSD of three BVP signals extracted from three facial regions exhibit the same peak and share the close frequency band with dense power during the same period, the periodicity and morphology still show notable differences. Similarly, under the TTA process, the strong regularization of frequency-domain consistency will force the model to capture only the average level of facial color changes among all regions. It might work in supervised learning or achieving generalization, but it is insufficient to satisfy the need to adapt to a specific user or scenario in TTA.

\section{Methodology}

In this paper, we propose an expert knowledge-based self-supervised Consistency-inConsistency-integration (CiCi) framework to enhance TTA in RPM tasks. Firstly, we formulate our research question in Sec. \ref{subsection: Problem Formulation}. Next, we describe the preprocessing methods for mmWave signals and RGB videos in Sec. \ref{subsection: preprocess}. Then, we describe the specific design of STFC and STTI in Sec. \ref{subsection: Domain Knowledge-based Priors}. Then, our gradient control strategy GDC will be introduced in Sec. \ref{subsection: 3.4}

\subsection{Problem Formulation}
\label{subsection: Problem Formulation}

\begin{figure}[ht] 
    \centering
    \includegraphics[width=0.49\textwidth]{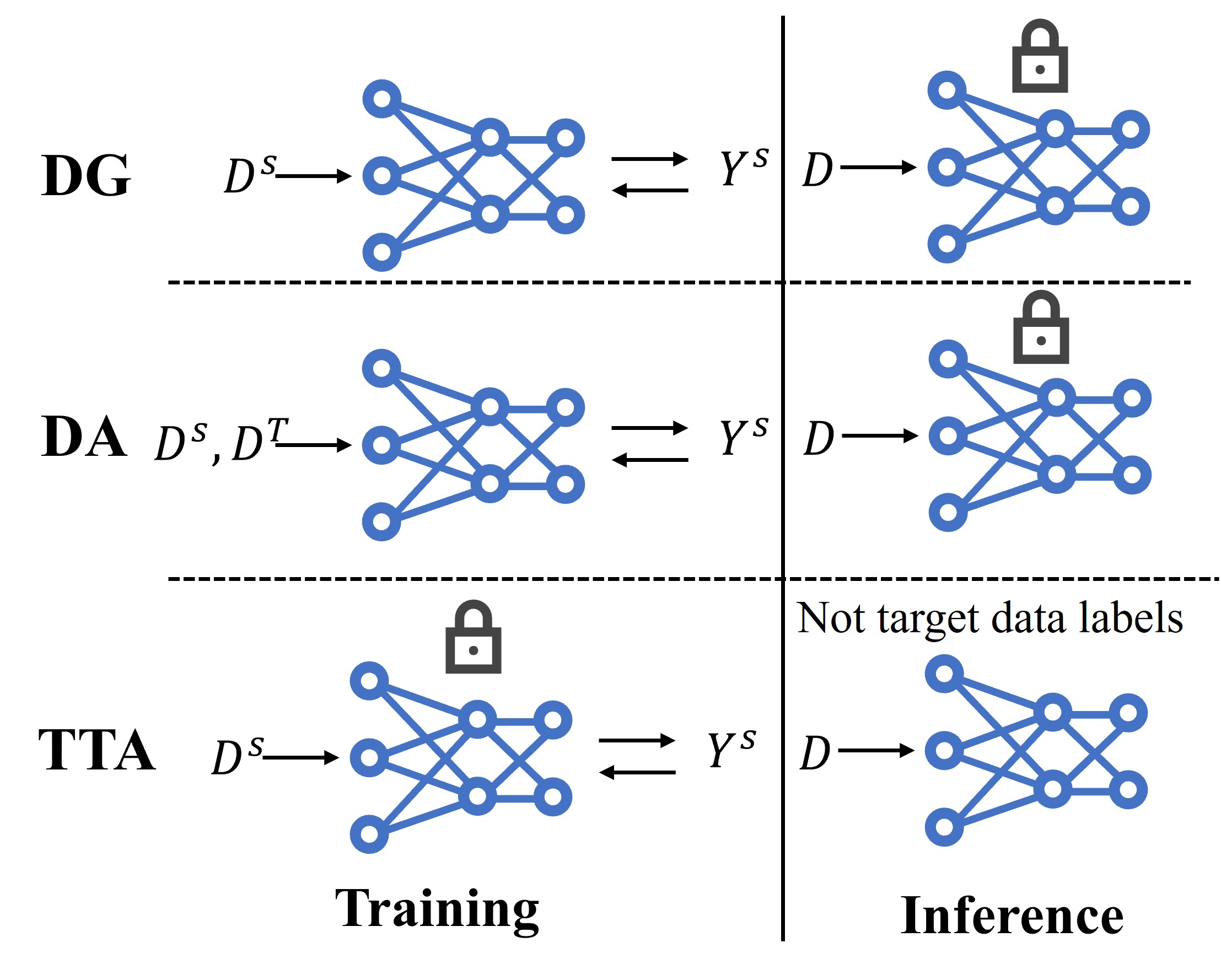} 
    \caption{Visualization of Domain Generalization (DG), Domain Adaptation (DA), and Test-Time Adaptation (TTA) methods. Here, $D^S$ represents the source domain data, $D^T$ represents the target domain data, and $Y^S$ represents the label of the source domain data. Notably, unlike DG and DA, TTA does not involve an offline training phase; instead, it performs unsupervised model adaptation during inference.}
    \label{fig:TTA-adv}
\end{figure}

Being different from DG and DA tasks, we do not focus on the training on the source domains $\mathcal{D}^S = \{X_i^S\}_{i=1}^N$. Instead, as shown in Fig. \ref{fig:TTA-adv}, the aim is to promote the performance of pre-trained models on an unlabeled target domain $\mathcal{D}^T = \{X_i\}_{i=1}^N$ by optimizing model parameters during inferring. The goal of this paper is to optimize and adapt the model parameters $\theta$ to the most proper $\theta'$ in the target domain without supervision.

\subsection{Preprocess}
\label{subsection: preprocess}

\subsubsection{Spatio-temporal Map}
\begin{figure}[ht] 
    \centering
    \includegraphics[width=0.48\textwidth]{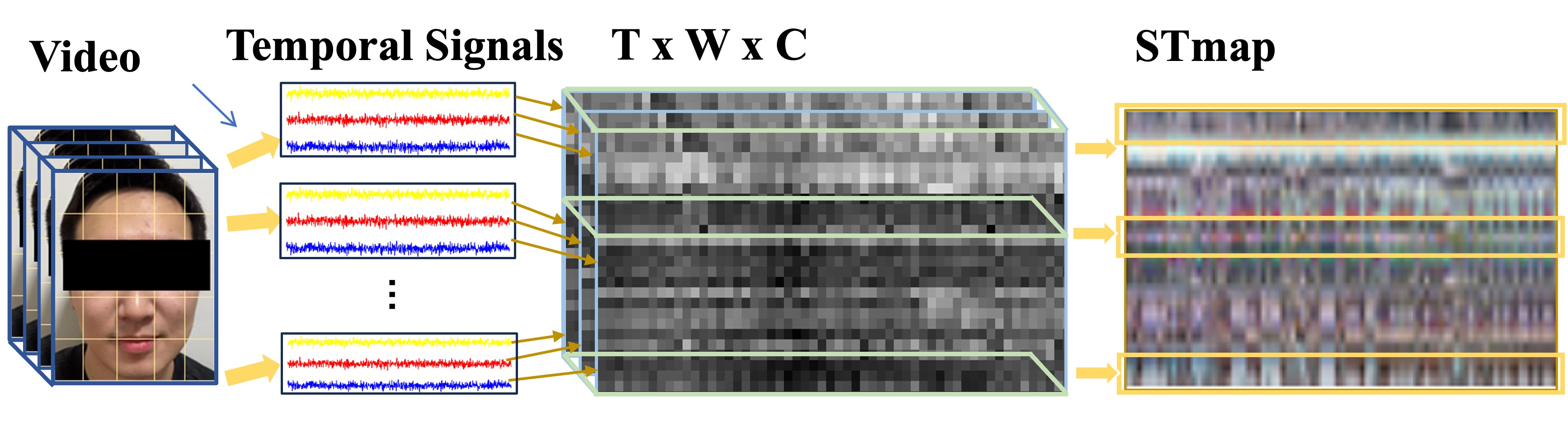} 
    \caption{The process of generating an STMap from video.}
    \label{fig:STMap}
\end{figure}

To reduce the computational resources consumed by video processing, inspired by \cite{niu2019rhythmnet}, we introduce STMap, which extracts the average pixel points of several skin regions of interest (ROIs) from each frame and combines them into a 2D physiological map that contains both spatial and temporal information. Specifically, the STMap $X_i \in \mathbb{R}^{T \times W \times 3}$ was preprocessed based on RGB facial video. Process procedure is shown in Fig. \ref{fig:STMap}, where the $T$ represents the number of frames, $W$ represents the average pixel values for each ROI on the face, and $3$ is the channel number of RGB figure.

\begin{figure}[ht] 
    \centering
    \includegraphics[width=0.48\textwidth]{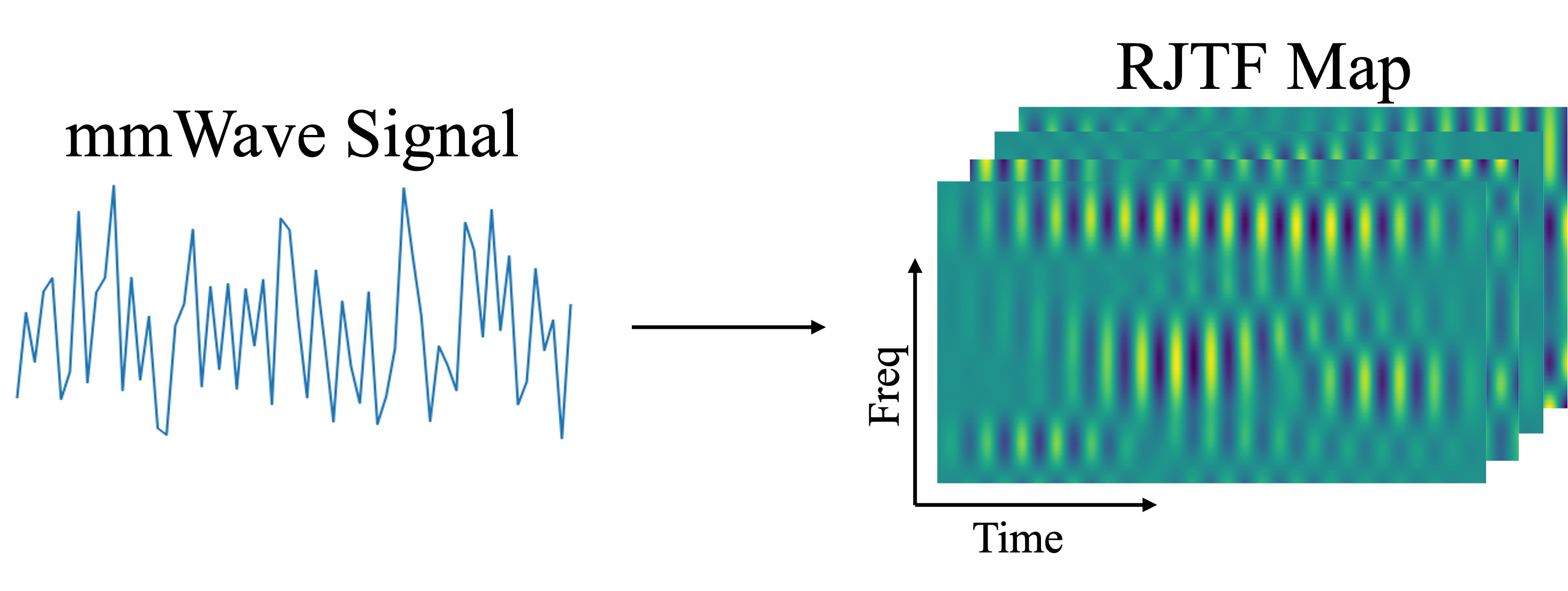} 
    \caption{The process of generating an RJTF map from mmWave signal.}
    \label{fig:RJTF}
\end{figure}

\subsubsection{Radio Joint Time-frequency Map}
The mmWave signals are often affected by multipath effects, noise, and environmental interference. To effectively extract useful information from the mmWave signals, the method described in \cite{choi2022remote} is applied for signal processing. Specifically, the Doppler effect serves as a key feature for analyzing moving targets within the signal. To capture the spatio-temporal characteristics of the signal, a Short-Time Fourier Transform is employed for time-frequency analysis, allowing the calculation of frequency shifts to extract Doppler features. Ultimately, this results in a radio joint time-frequency (RJTF) map $X_i \in \mathbb{R}^{T \times N_r \times F}$ of the signal, as shown in Fig. \ref{fig:RJTF}, where $N_r$ is the number of receiver channels, $F$ is the frequency resolution.

\subsection{Domain Knowledge-Based Priors}
\label{subsection: Domain Knowledge-based Priors}

\subsubsection{Spatio-temporal Augmentation}
\begin{figure}[ht] 
    \centering
    \includegraphics[width=0.4\textwidth]{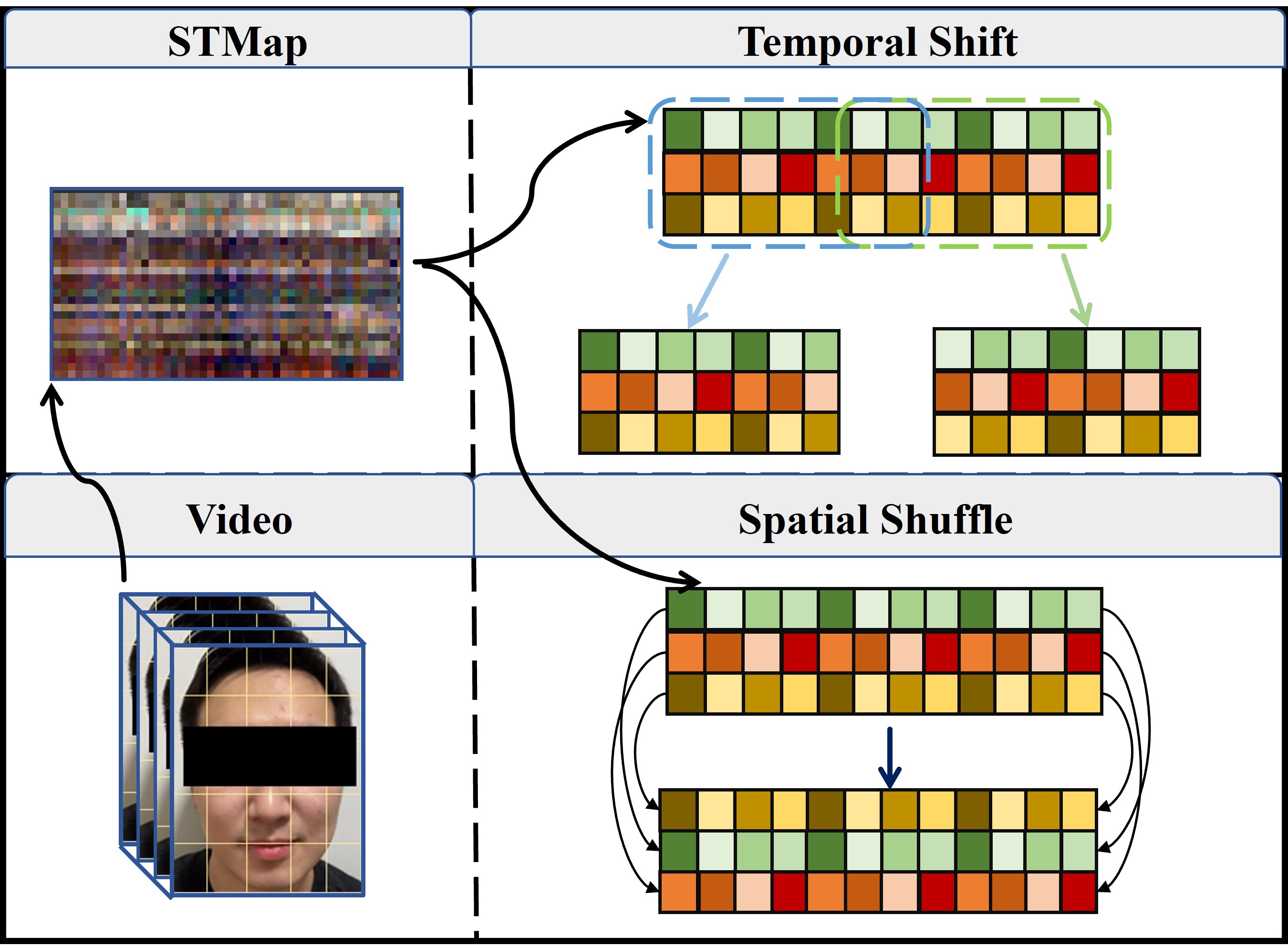} 
    \caption{Illustration of data augmentation on STMaps. In the temporal dimension, it randomly shifts forward by 0-30 frames during sliding segmentation. In the spatial dimension, it randomly scrambles the positions of pixels within the same frame}
    \label{fig:spatio-temporal augmentation}
\end{figure}

To construct the proper contrast identity for self-supervised learning, following previous work \cite{lu2023neuron,wang2025physmle}, we employ an image augmentation strategy (as illustrated in Fig. \ref{fig:spatio-temporal augmentation}) to construct augmentations with different spatial-temporal information. Specifically, we simulate irregular changes in spatial information by randomly permuting pixels along the $W$ dimension. This operation can be defined as performing an independent random permutation of columns for each time frame in the STMap and RJTF map, formally represented as permuting pixel $x_{t, i,j}$ at each time step $t$, where $i$ is the index on the $W$ dimension and $j$ represents the color channel. The permutation function is denoted as $\pi$ as $x_{\tau,\pi(i),j} \leftarrow x_{\tau,i,j}$, where $\pi$ is a random permutation of the set $\{1, 2, ..., W\}$.

In the temporal aspect, we introduce a random offset variable $\delta_t \sim Uniform(0,1,2,...,30)$ frame following a discrete uniform distribution to simulate potential time delays and sampling inconsistencies in physiological signal acquisition. The $\delta_t$ is applied to the $T$ dimension of $X$ as $x_{\tau,i,j} \leftarrow x_{\tau+\delta_t,i,j}$. By introducing spatio-temporal enhancement mechanisms in both temporal and spatial dimensions, the augmented STMap $X^a$ undergoes changes in the BVP signals, compared to the original STMap's BVP signals. The augmented input $X^a$ and the original input $X$ are utilized for self-supervised learning.

\subsubsection{Spatio-temporal Frequency-domain Consistency Regularization}
To align the frequencies of sharp peaks in the PSD of BVP signals corresponding to $X_i$ and $X_i^{a}$, we first calculate the shifts in the peak points of the PSD between the augmented and original signals as $\Delta P(f) = \lvert P_{a}(f) - P(f) \rvert$, where $P_{a}(f)$ and $P(f)$ denote the frequency at which the PSDs sharp peaks of the augmented and original signals are located. In order to filter out small irrelevant fluctuations but improve sensitivity to significant changes, we apply a characteristic function $\chi_{\psi}(f)$, defined by the threshold $\psi$, to filter the PSD differences:

\begin{equation}
\chi_{\psi}(f) = \begin{cases} 
1 & \text{if } \Delta P(f) \geq \psi, \\
0 & \text{otherwise}.
\end{cases}
\end{equation}

This selective approach ensures that our L1 loss calculation includes only those frequency variations that surpass $\psi$, thus focusing on differences of physiological importance. The filtered differences $\Delta P_{filterd}(f) = \Delta P(f).\chi_{\psi}(f) $ are then applied to compute the STFC loss $\mathcal{L}_{\text{STFC}}$ across the entire frequency domain:

\begin{equation}
\mathcal{L}_{\text{STFC}} = \frac{1}{N} \sum_{i=1}^{N} \left| \Delta P_{\text{filtered}}(f) \right|
\end{equation}

This soft regularization term can help the model only focus on significant errors, thereby reducing the impact of minor bias or deviations.

\subsubsection{Spatio-temporal Time-domain Inconsistency Regularization}

As mentioned in Sec. \ref{section: Observations About rPPG}, solely regularizing with consistency cannot achieve the adaptation to specific individual or deployment scenarios. Thus, combing the variations of BVP signals after spatio-temporal augmentation in the amplitude and shape, we designed the STTI regularization to assist the model in fitting the target domain by paying attention to the detailed semantic differences.

\begin{figure}[ht] 
    \centering
    \includegraphics[width=0.45\textwidth]{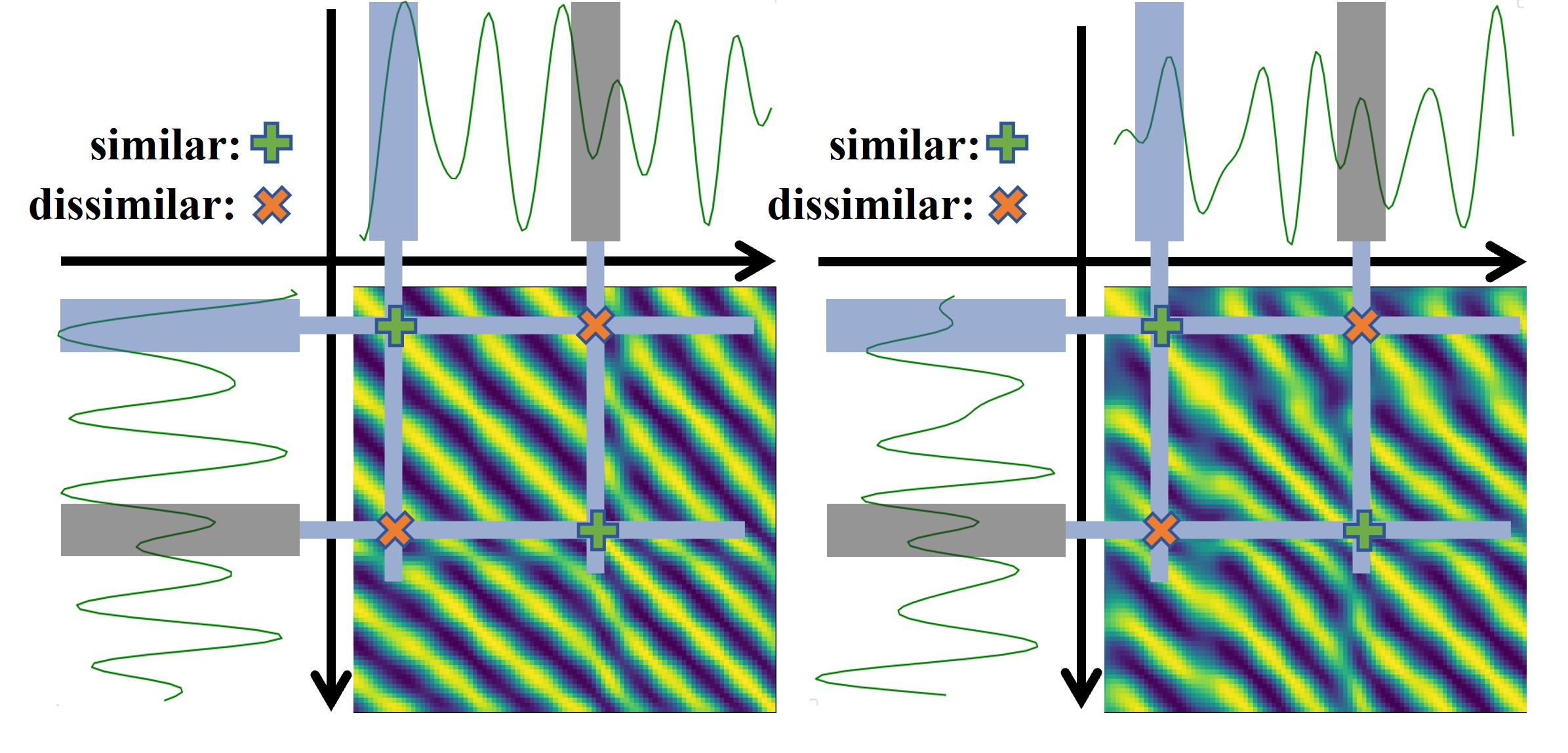} 
    \caption{Illustration of the self-similarity matrix generated from the BVP signals of the STMap before and after data augmentation. The brightness in the image indicates the degree of temporal information similarity between the windows.}
    \label{fig:ssm}
\end{figure}

The BVP signals generated by inputs before and after data augmentation exhibit a phenomenon of phase delay. Such delays can lead the model to excessively focus on the differences caused by the time delay when using STTI, thereby neglecting differences in waveform amplitude and shape. Inspired by \cite{sun2023resolve}, we use a self-similarity matrix (SSM) to mitigate the impact of phase delay. 

\begin{figure}[ht] 
    \centering
    \includegraphics[width=0.45\textwidth]{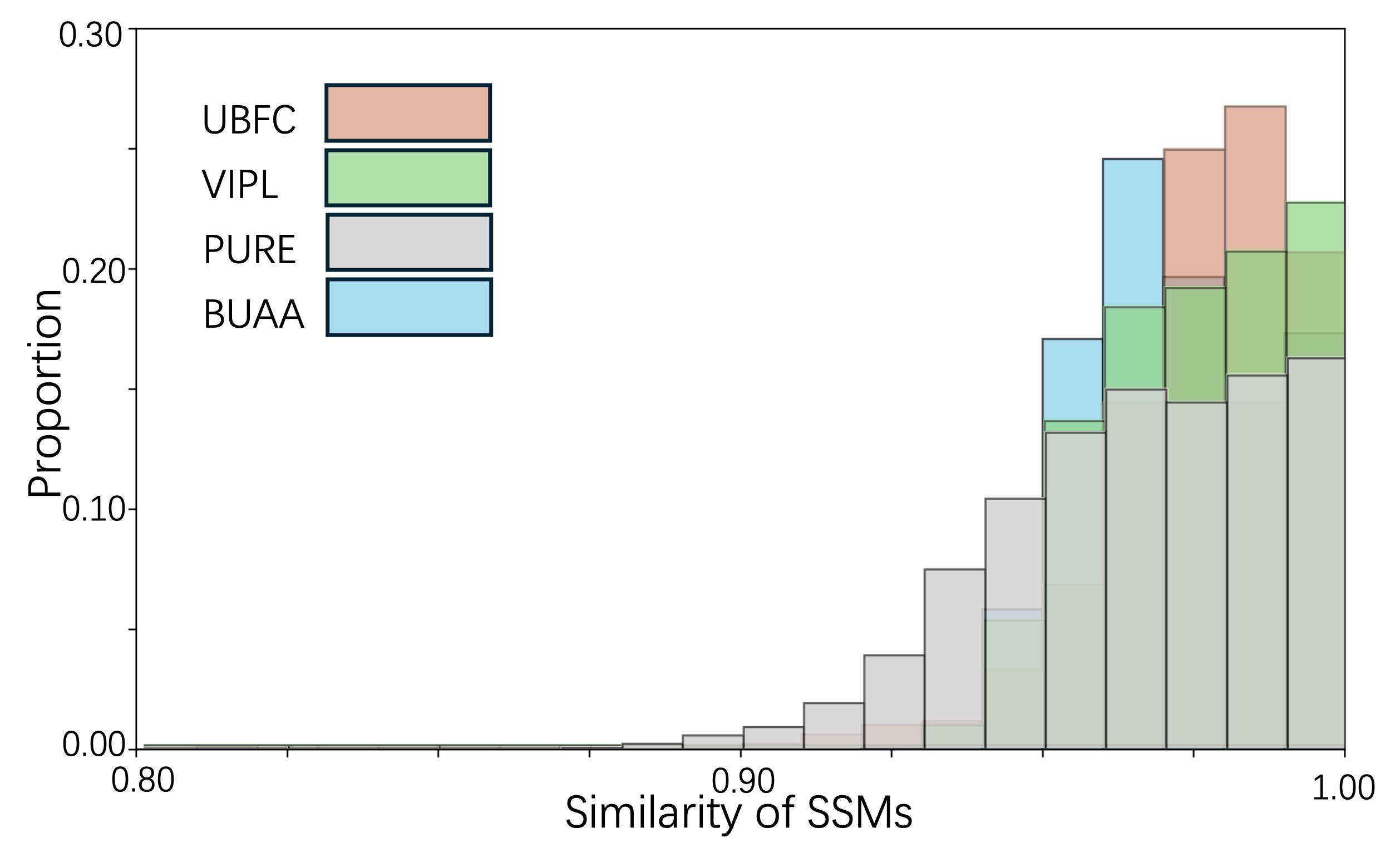} 
    \caption{The visualized spatio-temporal inconsistency in different datasets}
    \label{fig:Similarity of SSMs}
\end{figure}

Specifically, we use a sliding window of length $s$, with a step size of $1$, to extract the time windows $U = \{u_1,u_2,u_3,...,u_{T-s+1}\}$ from the BVP signals generated by the STMap. By iterating through all elements in $U$ and calculating the cosine similarity between each pair of elements 
$(u_i, u_j)$, we construct the following self-similarity matrix 
$\mathcal{M} \in \mathbb{R}^{(T-T_i+1)\times(T-T_i+1)}$:

\begin{equation}
\mathcal{M}_{ij} = Sim(u_i,u_j) = \frac{u_i \cdot u_j}{||u_i||_2 \cdot ||u_j||_2}
\end{equation}
Referring to Fig. \ref{fig:ssm}, even when two BVP signals reflect identical heart rates, they can exhibit distinct morphological characteristics due to variations in spatial dimensions, which manifests as subtle but meaningful differences in their respective similarity matrices and can lead to variations in the values at each matrix point of the self-similarity matrix, during the process of iterating over different time windows to calculate cosine similarities. These variations manifest as differences in local feature representations across different similarity matrices. To verify the existence of spatiotemporal inconsistency, we statistically analyzed the similarity between the SSMs corresponding to each segment of the original BVP signals and their augmented counterparts across different datasets (UBFC-rPPG \cite{de2013robust}, VIPL \cite{lu2021dual}, PURE \cite{song2021pulsegan}, BUAA \cite{dong2024realistic}, which will be introduced later in Sec. \ref{subsection: Datasets and Baselines}), as shown in Fig. \ref{fig:Similarity of SSMs}. Here, a higher similarity indicates smaller spatio-temporal differences between the two signals, while a lower similarity reflects greater discrepancies. It is worth noting that although the similarity tends to approach 1, each signal still exhibits a certain degree of variation due to external environmental factors, device latency, and physiological fluctuations in the human body. This variation is what we define as spatio-temporal inconsistency.

Therefore, to capture these variations in local features, we introduce a $\mathcal{L}_{STTI}$ loss that explicitly optimizes for the dissimilarity between flattened similarity matrices of original BVP signals, where $\mathcal{M}_{flat}$ corresponds to augmented BVP signals $\mathcal{M}_{flat}^{a}$:
\begin{equation}
    \mathcal{L}_{\text{STTI}} = 
    \frac{1}{N}\sum^N \frac{\mathcal{M}_{\text{flat}}\cdot \mathcal{M}_{\text{flat}}^{a}}{\|\mathcal{M}_{\text{flat}}\|\cdot\|\mathcal{M}_{\text{flat}}^{a}\|}
\end{equation}

By minimizing this loss term, our model is encouraged to learn and maintain the distinctive morphological characteristics between different BVP signals, rather than averaging or smoothing out these crucial variations. This approach ensures that the model captures the subtle yet important differences in signal patterns that arise from spatial variations, leading to more accurate and robust physiological measurements in the target domain.
\subsection{Gradient Dynamic Control (GDC)}
\label{subsection: 3.4}

As mentioned in the Sec. \ref{subsection: Domain Knowledge-based Priors}, we have employed $\mathcal{L}_{STFC}$ and $\mathcal{L}_{STTI}$ as our priors in the self-supervised tasks to balance generalizable invariant semantics and morphological differences under specific domains. Specifically, the frequency-domain consistency constrains ($\mathcal{L}_{STFC}$) the model to prevent rapid fluctuations in HR over short periods, while the temporal inconsistency ($\mathcal{L}_{STTI}$) encourages the model to fit the BVP signal towards high morphological variability. However, through mathematical analysis, we discovered that the gradients of these two loss functions exhibit significant optimization magnitudes differences in the later stages of training. 

Specifically, the consistency prior $\mathcal{L}_{STFC}$ is defined based on the absolute difference in peak frequency in PSD corresponding to the augmented and original input. Its gradient to model parameters $\theta$ is given by:

\begin{equation}
    \nabla_{\theta} \mathcal{L}_{\text{STFC}} = \frac{1}{N} \sum_{i=1}^{N} \text{sign}(\Delta P_{\text{filtered}}(f)) \cdot \chi_{\psi}(f)
\end{equation}

In the early stages of training when the difference is relatively large, the gradient $\nabla_{\theta} \mathcal{L}_{\text{STFC}}$ is also large, effectively driving the model to improve consistency. As the model is optimized, the gap narrows, and the gradient gradually decreases. However, as discussed in \cite{foret2020sharpness}, minimizing the loss function alone is insufficient to achieve good generalization. Additionally, as validated in \cite{li2024bi}, overfitting in the target domain model is inevitable. Therefore, after multiple training iterations, when the model begins to degrade, the HR deviation will widen again, causing the gradient of $\nabla_{\theta} \mathcal{L}_{\text{STFC}}$ to rise, ultimately leading the model to accelerate in an uncontrollable direction. In contrast, the gradients of $\mathcal{L}_{\text{STTI}}$ is:

\begin{equation}
\begin{aligned}
    \nabla_{\theta} \mathcal{L}_{\text{STTI}} = \frac{1}{N} \sum^N \left( \frac{\mathcal{M}_{\text{flat}}^{a}}{\|\mathcal{M}_{\text{flat}}\| \cdot \|\mathcal{M}_{\text{flat}}^{a}\|} - \frac{\mathcal{M}_{\text{flat}} \cdot \mathcal{M}_{\text{flat}} \cdot\mathcal{M}_{\text{flat}}^{a} }{\|\mathcal{M}_{\text{flat}}\|^3 \cdot \|\mathcal{M}_{\text{flat}}^{a}\|^2}  \right)
\end{aligned}
\end{equation}

The purpose of $\mathcal{L}_{\text{STTI}}$ is to encourage the self-similarity matrix of BVP signals to evolve toward a state of inconsistency. 
Thus, in the early stages of training, influenced by the strong generalization capability of the pretrained model, the differences between 
$\mathcal{M}_{\text{flat}}$ and $\mathcal{M}_{\text{flat}}^{a}$ are relatively small. This results in a smaller 
$\mathcal{M}_{\text{flat}} \cdot \mathcal{M}_{\text{flat}}^{a}$ and larger $L2$-norms, which in turn leads to weaker gradients. After multiple iterations, the model begins to adapt to the specific characteristics of the dataset, and the dissimilarity between $\mathcal{M}_{\text{flat}}$ and $\mathcal{M}_{\text{flat}}^{a}$ becomes more pronounced, causing the dot product to increase along with the gradient magnitude.

\begin{figure}[ht] 
\vspace{-3mm}
    \centering
    \includegraphics[width=0.48\textwidth]{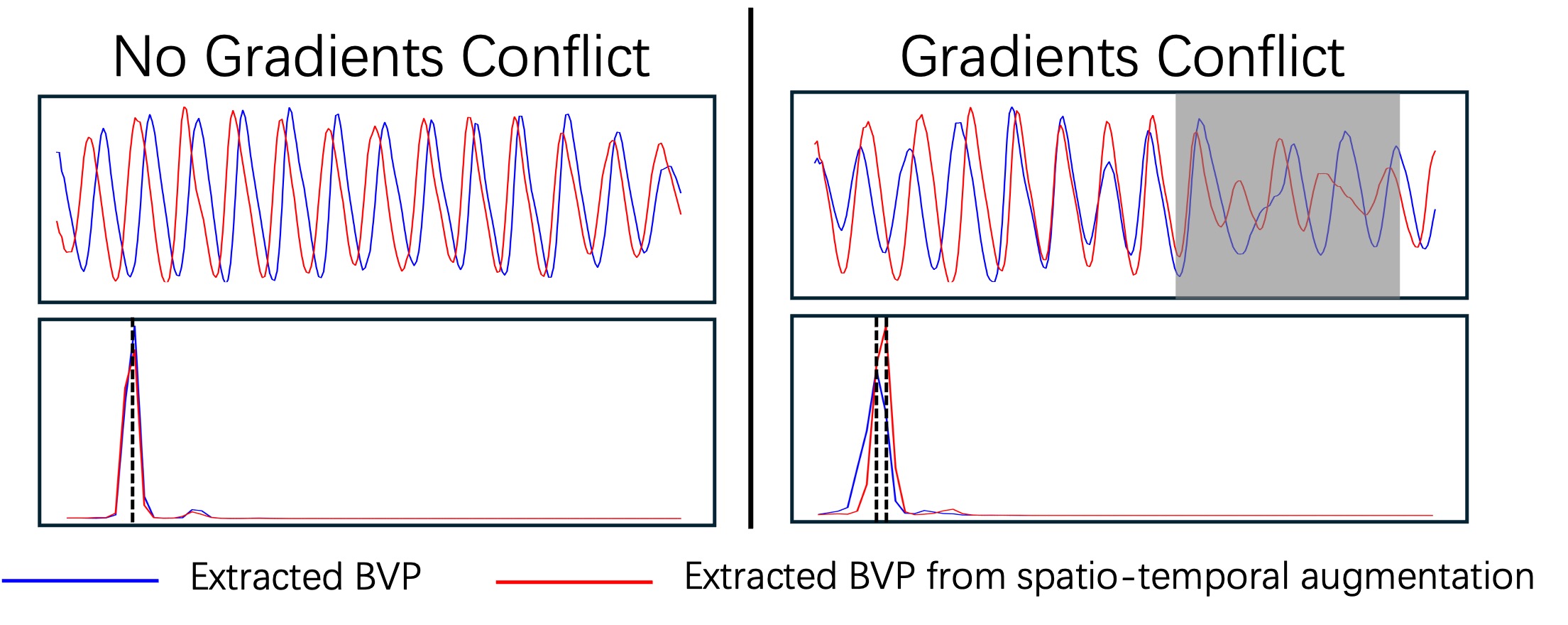} 
    \vspace{-3mm}
    \caption{Visualization the difference between the time-domain and the frequency-domain when the gradients conflict}
    \vspace{-3mm}
    \label{fig:gradients conflict}
\end{figure}

Simultaneously, we analyzed the directions of optimization, which can be categorized into four scenarios: (1) The predicted BVP signal is highly consistent in the frequency domain and has large temporal variations, which is the ideal case with minimal model updates; (2) The frequency domain is highly inconsistent, and temporal variations are small, indicating that the output BVP is highly unreliable, prompting the model to learn towards frequency-domain consistency; (3) The frequency domain is highly consistent, but temporal variation is small, neglecting morphological variability, causing the model to learn towards temporal inconsistency; (4) The frequency domain is highly inconsistent, and temporal variation is large. In this case, the model is simultaneously influenced by two competing gradients. As a result, it attempts to both amplify morphological differences and improve frequency domain consistency. However, the excessive morphological discrepancies cause the BVP signal’s intrinsic features to be lost, making it increasingly difficult to enhance frequency domain consistency, as shown in Fig. \ref{fig:gradients conflict}. This leads to a conflict between the two gradients (defined as $\nabla_{\mathcal{L}_{\text{STFC}}} \cdot \nabla_{\theta} \mathcal{L}_{\text{STTI}} < 0$). Furthermore, analysis of the gradient magnitude trends shows that in such cases, both gradients remain relatively large, with neither dominating the averaged gradient. Consequently, the model tends to generalize in an uncontrollable direction.

To prevent model collapse caused by gradient conflicts, we designed a Gradient Dynamic Control (GDC) strategy to adaptively balance these two conflicting gradients. Specifically, when the dot product of the two gradients is negative, we consider the gradients to be in conflict, we prioritize $\mathcal{L}_{\text{STFC}}$, encouraging the model to focus on frequency domain consistency first and thereby avoid over-reliance on the inconsistency prior that could lead to collapse. Inspired by \cite{yu2020gradient}, when the $\mathcal{L}_{\text{STFC}}$ and $\mathcal{L}_{\text{STTI}}$ gradients conflict, we calculate the gradient  $\nabla_{\theta} \mathcal{L}_{\text{STTI}}$ and $\nabla_{\theta} \mathcal{L}_{\text{STFC}}$, and project $\nabla_{\theta} \mathcal{L}_{\text{STTI}}$ onto the orthogonal direction of the $\nabla_{\theta} \mathcal{L}_{\text{STTI}}$ to avoid from undermining $\nabla_{\theta} \mathcal{L}_{\text{STFC}}$. The formulation is as follows:

\begin{equation}
    \nabla_{\theta} \mathcal{L} = \nabla_{\theta} \mathcal{L}_{\text{STFC}} - \frac{\nabla_{\theta} \mathcal{L}_{\text{STFC}} \cdot \nabla_{\theta} \mathcal{L}_{\text{STTI}}}{\|\nabla_{\theta} \mathcal{L}_{\text{STTI}}\|^2} \cdot \nabla_{\theta} \mathcal{L}_{\text{STTI}}
\end{equation}

In this way, we update the combined gradient $\nabla_{\theta} \mathcal{L}$ to balance the mutual influence of the two gradients when there is conflict, leading to more stable adapting. When there is no conflict, our optimization goal is $\nabla_{\theta} \mathcal{L} = \nabla_{\theta} \mathcal{L}_{STFC} + \nabla_{\theta} \mathcal{L}_{STTI}$.

\begin{table*}[ht]
\setlength{\tabcolsep}{1mm}
\centering
\scriptsize
\caption{HR estimation results on MSDG and TTA protocol. For better comparison, parts of the results in this table are from \cite{li2024bi}. In this and the following tables, the best results for given metrics are bolded, the second-best is underlined. The '\(^*\)' shows the significantly (p-value $<$ 0.05) best result within each column with the paired-t test.}

\begin{tabular}{lccccccccccccccc}
\toprule
  &\multicolumn{3}{c}{\textbf{UBFC}} & \multicolumn{3}{c}{\textbf{PURE}} & \multicolumn{3}{c}{\textbf{BUAA}} & \multicolumn{3}{c}{\textbf{VIPL}} & \multicolumn{3}{c}{\textbf{V4V}} \\
  \cmidrule(lr){2-4} \cmidrule(lr){5-7} 
\cmidrule(lr){8-10} \cmidrule(lr){11-13} \cmidrule(lr){14-16} 
\textbf{Method} & \textbf{MAE↓}  & \textbf{RMSE↓} & \textbf{\quad p↑ \quad}    & \textbf{MAE↓}  & \textbf{RMSE↓} & \textbf{\quad p↑ \quad}    & \textbf{MAE↓}  & \textbf{RMSE↓} & \textbf{\quad p↑ \quad}    & \textbf{MAE↓}  & \textbf{RMSE↓}  & \textbf{\quad p↑ \quad} & \textbf{MAE↓}  & \textbf{RMSE↓}  & \textbf{\quad p↑ \quad}    \\
\midrule
\multicolumn{16}{c}{\textbf{Methods based on MSDG protocol}} \\
\midrule

GREEN \cite{verkruysse2008remote} & 8.02 & 9.18 & 0.36 & 10.32 & 14.27 & 0.52 &5.82 &7.99 &0.56 & 12.18 & 18.23 & 0.25 & 15.64 & 21.43 & 0.06 \\
CHROM \cite{de2013robust} & 7.23 & 8.92 & 0.51 & 9.79 & 12.76 & 0.37 &6.09 &8.29 &0.51 & 11.44 & 16.97 & 0.28 & 14.92 & 19.22 & 0.08 \\
POS \cite{wang2016algorithmic} & 7.35 & 8.04 & 0.49 & 9.82 & 13.44 & 0.34 &5.04 &7.12 &0.63 & 14.59 & 21.26 & 0.19 & 17.65 & 23.22 & 0.04 \\
\midrule
DeepPhys \cite{chen2018deepphys} & 7.82 & 8.42 & 0.54 & 9.34 & 12.56 & 0.55 &4.78 &6.74 &0.69 & 12.56 & 19.13 & 0.14 & 14.52 & 19.11 & 0.14 \\
TS-CAN \cite{liu2020multi} & 7.63 & 8.25 & 0.55 & 9.12 & 12.38 & 0.57 &4.84 &6.89 &0.68 & 12.34 & 18.94 & 0.16 & 14.77 & 19.96 & 0.12 \\
RhythmNet \cite{niu2019rhythmnet} & 5.79 & 7.91 & 0.78 & 7.39 & 10.49 & 0.77 &3.38 &5.17 &0.84 & 8.97 & 12.16 & 0.49 & 10.16 & 14.57 & 0.34 \\

PulseGAN \cite{song2021pulsegan} & 5.61 & 7.83 & 0.78 & 7.31 & 10.40 & 0.78 &3.35 &5.19 &0.84 & 8.91 & 12.18 & 0.48 & 10.19 & 14.55 & 0.34 \\
Dual-GAN \cite{lu2021dual} & 5.55 & 7.62 & 0.79 & 7.24 & 10.27 & 0.78  & 3.41 & 5.23 & 0.84 & 8.88 & 11.69 & 0.50 & 10.04 & 14.44 & 0.35  \\
BVPNet \cite{das2021bvpnet} & 5.43 & 7.71 & 0.80 & 7.23 & 10.25 & 0.78 &3.69 &5.48 &0.81 & 8.45 & 11.64 & 0.51 & 10.01 & 14.35 & 0.36 \\
Physformer++ \cite{yu2023physformer} & 5.24 & 7.37 & 0.83 & 7.19 & 10.08 & 0.80 & 3.30 & 5.51 & 0.84 & 8.58 & 11.67 & 0.53 & 9.87 & 14.23 & 0.38\\
\midrule
AD \cite{ganin2015unsupervised} & 5.92 & 8.08 & 0.76 & 7.42 & 10.61 & 0.73 &3.49 &5.49 &0.82  & 8.41 & 11.71 & 0.53 & 10.47 & 14.64 & 0.32 \\
GroupDRO \cite{parascandolo2020learning} & 5.73 & 7.97 & 0.78 & 7.69 & 10.83 & 0.78 &3.41 &5.21 &0.83 &8.35 & 11.67 & 0.54 & 9.94 & 14.29 & 0.36 \\
Coral \cite{sun2016return} & 5.89 & 8.04 & 0.76 & 7.59 & 10.87 & 0.72 &3.64 &5.74 &0.80 & 8.68 &11.91 & 0.53 & 10.32 & 14.42 & 0.32 \\
Meta-rPPG \cite{lee2020meta} & 5.59 & 7.42 & 0.81 & 7.36 & 9.89 & 0.79 & 3.24 & 5.46 & 0.85 & 8.21 & 11.40 & 0.57 & 9.54 & 14.13 & 0.40 \\
VREx \cite{krueger2021out} & 5.59 & 7.68 & 0.81 & 7.24 & 10.14 & 0.78 &3.27 &5.01 &0.86 &8.47 &11.62 & 0.54 & 9.82 & 14.16 & 0.37 \\
NEST \cite{lu2023neuron} & 4.67 & 6.79 & 0.86 & 6.71 &9.59 & 0.81 &2.88 &4.69 &0.89 &7.86 &11.15 & 0.58 & 9.27 & 13.79 & 0.41 \\
HSRD \cite{wang2023hierarchical} & 4.31 & 6.30 & 0.88 & 6.35 & 9.36 & 0.82 & 2.71 & 3.09          & 0.90 & 8.00 & 11.20 & 0.57 & 8.66 & 13.51 & 0.43 \\
\midrule
\multicolumn{16}{c}{\textbf{Methods based on TTA protocol}} \\
\midrule 
TENT \cite{wang2020tent} & 4.48 & 9.33 & 0.88 & 6.77 & 9.61 & 0.85 & 2.31 & 3.41 & 0.93 & 7.92 & 10.99 & 0.55 & 9.71 & 11.69 & 0.47 \\
SAR \cite{niu2023towards} & 4.52 & 9.49 & \underline{0.86} & 6.34 & 9.26 & 0.86 & 2.02 & 2.93 & 0.94 & 7.97 & 11.46 & 0.54& 9.18 & 10.91 & 0.48\\
TTT++ \cite{liu2021ttt++} & 4.32 & 6.52 & 0.87 & 6.17 & 9.02 & 0.85 & 2.04 & 2.87 & 0.92 & 7.41 & 11.25 & 0.57 & 9.36 & 11.43 & \underline{0.53}\\
EATA \cite{niu2022efficient} & 4.04 & 6.43 & 0.89 & 6.02 & 8.85 & 0.87 & 1.72 & 2.43 & 0.95 & 7.30 & 11.01 & 0.57 & 8.42 & 11.52 & \underline{0.53} \\
SHOT \cite{liang2021source} & 4.19 & 6.41 & 0.87 & 5.81 & 8.52 & 0.88 & 1.77 & 2.32 & 0.97 & 7.54 & 10.92 & 0.57 & 8.71 & 11.29 & 0.49\\
DeYO \cite{leeentropy} & 3.97 & 6.36 & 0.87 & 5.93 & 8.42 & 0.89 & 1.68 & 2.28 & 0.96 & 7.31 & 10.90 & 0.59 & 8.59 & 11.19 & 0.51\\
Bi-TTA \cite{li2024bi}& \underline{3.62} & \underline{6.33} & 0.87 & \underline{5.21} & \underline{8.38} & \underline{0.89} & \underline{1.63} & \underline{2.01} & \underline{0.99} & \underline{7.24} & \underline{10.88} & \underline{0.63} & \underline{8.61} & \underline{11.09} & 0.52   \\ 
\midrule
\textbf{CiCi}  & \textbf{3.51}\(^*\) & \textbf{6.18}\(^*\) & \textbf{0.90} & \textbf{4.87}\(^*\) & \textbf{7.92}\(^*\) & \textbf{0.91}\(^*\) & \textbf{1.53}\(^*\) & \textbf{1.99}\(^*\) & \textbf{0.99} & \textbf{6.84}\(^*\) & \textbf{10.52}\(^*\) & \textbf{0.66}\(^*\) & \textbf{8.56}\(^*\) & \textbf{10.97}\(^*\) & \textbf{0.55}\(^*\)\\ 
\bottomrule

\end{tabular}
\label{tabel: HR Estimation}
\end{table*}

\section{Experiments}

\subsection{Datasets and Baselines}
\label{subsection: Datasets and Baselines}

In this study, we focused on five distinct RGB datasets that encompass a broad spectrum of motion dynamics, camera configurations, and lighting conditions. Additionally, we also selected two different mmWave datasets that include different collection conditions. 

Specifically, {\bfseries UBFC-rPPG} \cite{de2013robust} consists of 42 facial videos captured under sunlight and indoor lighting conditions. {\bfseries PURE} \cite{song2021pulsegan}
comprises 60 RGB videos featuring 10 subjects engaged in six distinct activities. {\bfseries VIPL} \cite{lu2021dual}
includes nine distinct scenarios filmed using three RGB cameras under varying illumination conditions and different levels of motion. {\bfseries BUAA} \cite{dong2024realistic}
is specifically designed to assess the performance of algorithms across different illumination levels. {\bfseries V4V} \cite{li2024bi}
is specifically designed to capture data that exhibits significant fluctuations in physiological indicators across ten simulated tasks. {\bfseries EquiPleth} \cite{vilesov2022blending} contains FMCW radar data and RGB videos from 91 experimental subjects, with a radar starting frequency of 77 Hz. Each subject has been recorded across 6 different scenes, each lasting 30 seconds. {\bfseries PhysDrive} \cite{wang2025physdrive} includes data from 48 drivers under 3 different driving conditions and 4 different weather scenarios, along with physiological signals. It also contains multimodal data, such as mmWave radar, RGB cameras, and near-infrared cameras.

Notably, the BVP signals and video data within VIPL and V4V are not temporally synchronized. Thus, we did not evaluate the BVP and HRV estimation performance over it. Additionally, to ensure accurate alignment with the video data, the BVP signals in VIPL and V4V were down-sampled from 60 frames per second (fps) to 30 fps using cubic spline interpolation.

To compare the performance of CiCi across different datasets, we evaluated it under MSDG \cite{lu2023neuron} and TTA protocols \cite{li2024bi}. We chose traditional methods (i.e., GREEN \cite{verkruysse2008remote}, CHROM \cite{de2013robust}, POS \cite{wang2016algorithmic} ), deep learning-based methods (i.e., DeepPhys \cite{chen2018deepphys}, TS-CAN \cite{liu2020multi}, RhythmNet \cite{niu2019rhythmnet}, PulseGAN \cite{song2021pulsegan}, Dual-GAN \cite{lu2021dual}, BVPNet \cite{das2021bvpnet},   Physformer++ \cite{yu2023physformer}), DG-based methods (i.e., AD \cite{ganin2015unsupervised}, GroupDRO \cite{parascandolo2020learning}, Coral \cite{sun2016return}, VREx \cite{krueger2021out}, Meta-rPPG \cite{lee2020meta}, NEST \cite{lu2023neuron}, HSRD \cite{wang2023hierarchical} ) and TTA-based methods (i.e., TENT \cite{wang2020tent}, SAR \cite{niu2023towards}, TTT++ \cite{liu2021ttt++}, EATA \cite{niu2022efficient}, SHOT \cite{liang2021source}, DeYO \cite{leeentropy}, Bi-TTA \cite{li2024bi} ) to assess the performance of various approaches from the perspectives of mean absolute error (MAE), root mean square error (RMSE), and Pearson’s correlation coefficient (p). Additionally, five evaluations were initialized with five different random seeds, with a paired t-test applied to examine the significance of the average performance difference between the best-performing model and the second-best-performing model. Note that, as we perform TTA based on the best DG method, the results of TTA baselines were not based on \cite{li2024bi} but were evaluated based on our own results. 


\subsection{Implementation Details}
Our model was implemented on the Pytorch framework. We followed the method in \cite{lu2023neuron} to generate the STMaps from frame images. Then the STMap $X \in \mathbb{R}^{256 \times 25 \times 3}$ was resized to $X \in \mathbb{R}^{256 \times 64 \times 3}$, and the RJFT map was resized to $X \in \mathbb{R}^{256 \times 64  \times 4}$. After spatio-temporal augmentation, $X$ and $X^a$ were used as input. 

For the backbone network in rPPG task, ResNet-18 \cite{he2016deep} served as the feature extractor for HR estimation, with the max pooling layer removed to better retain spatial details. The final fully connected layer was then repurposed as the HR estimation head. Moreover, the BVP signal estimation head comprised four blocks, each containing a convolutional layer, and a transposed convolutional layer, followed by two convolutional layers with BN and ReLU activation functions. For the mmWave task, the RF-vital \cite{choi2022remote} encoder was chosen for feature extraction. Then, a fully connected layer was connected as the output head for the BVP signal. 

We used the stochastic gradient descent (SGD) with a momentum of 0.9 as the base optimizer. In addition, in our experiment, we set hyper-parameters $\lambda$, $\psi$ to 0.01, 1. The learning rate was 0.0001 and the batch size was 1. Finally, all TTA methods were first trained using HSRD on the other four datasets before testing on the target domain.

\subsection{HR Estimation in rPPG}

According to the MSDG protocol \cite{lu2023neuron} and TTA protocol \cite{li2024bi} in rPPG task, our study used four different datasets as source domains to train the offline model, with another dataset serving as the target domain to evaluate the performance of our method. Tab. \ref{tabel: HR Estimation} showed that DL-based algorithms generally outperformed traditional methods across datasets, particularly on the VIPL and V4V datasets, which indicates the greater potential of DL methods in more complex scenarios. Besides, the results of specifically designed DG algorithms showed general improvement (e.g. NEST, HSRD) for better generalization. However, it is important to note that the performance of some TTA methods was worse (such as TENT and SAR) compared with some DG methods (e.g. NEST, HSRD), which might be attributed to the fact that these methods were originally developed for classification tasks, not regression tasks (here, we refer to the implementation of Bi-TTA, where HR values between 40 and 160 in the frequency domain are divided into 121 categories in order to convert the task into a classification problem).  At the same time, TTA methods specifically designed for rPPG tasks (such as Bi-TTA) performed well in rPPG tasks. Despite this, our proposed CiCi achieved better results than all other algorithms across these five datasets. Notably, the CiCi method showed a significant better performance on the VIPL dataset compared to that on other datasets, which might be due to the less stable frame rate and greater temporal inconsistencies on the VIPL dataset, making the STTI rules more effective in capturing specific characteristics. In contrast, on more stable datasets such as V4V, where temporal inconsistency is minimal and frequency domain consistency is more pronounced, gradient conflicts were more likely to be exist, leading to smaller performance gains. This indicates that CiCi can outperform other methods in real-world environments with more noticeable device instability.

\begin{table}[ht]
\setlength{\tabcolsep}{1mm}
\centering
\scriptsize
\caption{Ablation study of different settings.}
\setlength{\tabcolsep}{1.4pt}
\begin{tabular}{lccccccccc}
\toprule
  \multicolumn{3}{c}{\textbf{Method}} & \multicolumn{3}{c}{\textbf{VIPL}} & \multicolumn{3}{c}{\textbf{V4V}} \\
  \cmidrule(lr){4-6} \cmidrule(lr){7-9} 
\textbf{$\mathcal{L}_{\text{STFC}}$} & \textbf{$\mathcal{L}_{\text{STTI}}$} & GDC & \textbf{MAE↓}  & \textbf{RMSE↓} & \textbf{\quad p↑ \quad}    & \textbf{MAE↓}  & \textbf{RMSE↓} & \textbf{\quad p↑ \quad} \\
\midrule
- & - & - & 8.00 & 11.20  & 0.57 & 8.66 & 13.51 & 0.43\\
\checkmark & - & -  & 6.96 & 10.73 & 0.63 & 8.62 & \underline{11.27} & 0.53 \\
- & \checkmark & -  & 7.11 & 10.90 & 0.62 & 9.42 & 15.11 & 0.35 \\
\checkmark & \checkmark & - & \underline{7.01} & \underline{10.71} & \underline{0.64} & \underline{8.61} & 11.34 & \underline{0.54} \\
\checkmark & \checkmark & \checkmark  & \textbf{6.84}\(^*\) & \textbf{10.52}\(^*\) & \textbf{0.66}\(^*\) & \textbf{8.56}\(^*\) & \textbf{10.97}\(^*\) & \textbf{0.55}\(^*\)\\ 
\bottomrule
\end{tabular}
\label{tabel: Ablation study}
\end{table}

\subsection{Ablation Study}

\begin{figure}[ht] 
    \centering
    \includegraphics[width=0.45\textwidth]{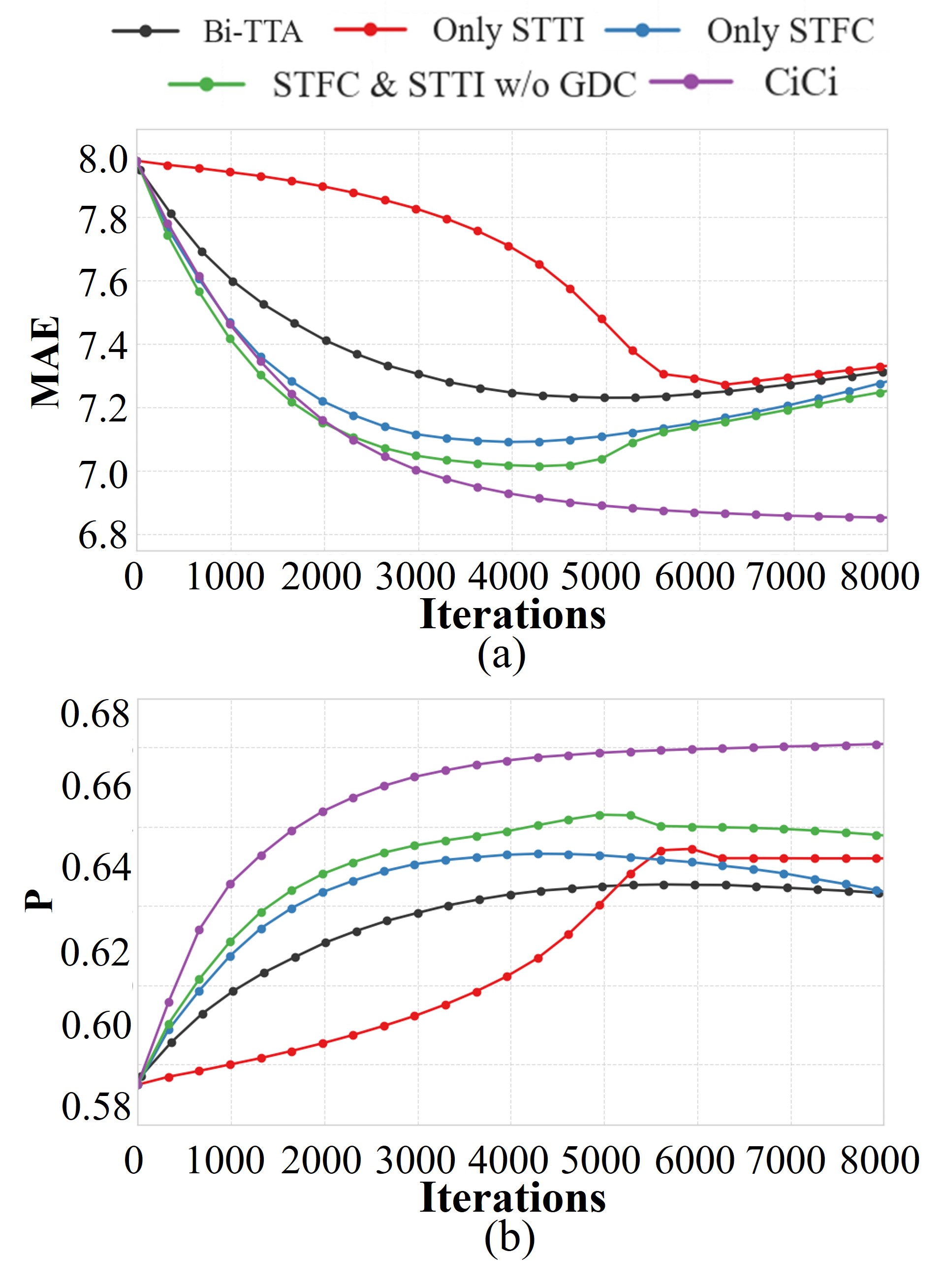} 
    \caption{The visualized performance of CiCi in the VIPL dataset in each iteration, which aims to present the impact of STTI, STFC, and GDC. The performance metrics are (a) MAE, (b) p.}
    \label{fig:ablation}
\end{figure}

In this section, we will discuss the effectiveness of ${L}_{\text{STTI}}$ and ${L}_{\text{STFC}}$, as well as GDC. From Tab. \ref{tabel: Ablation study}, we can observe that even when only using STFC as the sole prior condition, the model's performance showed a clear improvement compared to the pre-trained model. However, using only STTI as the loss function can lead to model collapse on some datasets. This may be due to that the inconsistency loss function in $\mathcal{L}_{STTI}$ could drive the model towards complete disorder, which is contrary to the objective laws of physiological signals. In contrast, the existence of a threshold, $\psi$, in $\mathcal{L}_{STFC}$, enabled the model to ignore trivial variations, thereby prevented it from overly focusing on consistency itself and consequently avoided overfitting that could degrade performance. At the same time, to verify the specific impact of the threshold $\psi$ on $\mathcal{L}_{STFC}$, we conducted an ablation study on $\psi$ while using $\mathcal{L}_{STFC}$ alone. The results are presented in Tab. \ref{table: psi}. It shows that the model achieved its best performance when $\psi = 1$, while too high or too low $\psi$ negatively affected the model performance. It must be noted that the optimal exact value of $\psi$ has yet to be fully explored; nevertheless, we recommended setting an appropriate threshold to prevent the model from being disturbed by minor variations.

\begin{table}[ht]
\setlength{\tabcolsep}{10pt}
\caption{Ablation Study for Different $\psi$}
\centering
\begin{tabular}{lccccccc}
\toprule
\textbf{$\psi$} & 0 & 0.5 & 1.0 & 1.5 & 2.0 & 2.5 \\
\midrule
\textbf{UBFC} & 3.76 & 3.68 & 3.63 & 3.65 & 3.67 & 3.73 \\
\textbf{PURE} & 5.30 & 5.19 & 5.09 & 5.13 & 5.14 & 5.22 \\
\textbf{BUAA} & 1.62 & 1.57 & 1.57 & 1.58 & 1.58 & 1.63 \\
\textbf{VIPL} & 7.12 & 6.97 & 6.96 & 7.03 & 7.09 & 7.18 \\
\textbf{V4V} & 8.81 & 8.73 & 8.62 & 8.65 & 8.68 & 8.70 \\
\bottomrule
\end{tabular}
\footnotesize{\textit{The MAE values are reported as the evaluation metric.}}
\label{table: psi}
\end{table}

Additionally, in Fig. \ref{fig:ablation}, it is noticeable that any method using $\mathcal{L}_{STFC}$ without gradient management led to a degradation of model performance, indicating that minimizing consistency loss functions could easily lead to over-optimization on the target domain, causing the model to overfit. For the method using only STTI, the model's performance improvement was relatively slow, gradually increasing with more iterations. This aligns with our observations in Section \ref{subsection: 3.4}, where the gradient magnitude slowly increased as the number of iterations increased. However, after a certain number of iterations, the model performance began to deteriorate.

The experimental results showed that STFC regulation directly and effectively constrained the model. However, if only STTI regulation was used to supervise the model, the training outcomes lacked stability, and the model tended to undergo reverse optimization. Yet, with the implementation of a gradient control strategy within the CiCi framework, the inherent uncertainties of the STTI prior were mitigated, potentially due to constraints imposed by STFC, which allowed the model to better leverage both types of priors for improved performance. Additionally, the reduction in gradient conflicts led to a remarkable improvement in the convergence speed when employing the GDC method.

\begin{table*}
\setlength{\tabcolsep}{5pt} 
\centering
\scriptsize
\caption{HRV analysis results on MSDG protocol and TTA protocol.}
\begin{tabular}{clcccccccccccc}
\toprule
 &   & \multicolumn{3}{c}{\textbf{LFnu}}   & \multicolumn{3}{c}{\textbf{HFnu}}   & \multicolumn{3}{c}{\textbf{LF/HF}} & \multicolumn{3}{c}{\textbf{HR-(bpm)}}     \\
\cmidrule(lr){3-5} \cmidrule(lr){6-8} 
\cmidrule(lr){9-11} \cmidrule(lr){12-14}        
{\textbf{Target}}&{\textbf{Method}}  & \textbf{MAE↓}    & \textbf{RMSE↓}   & \textbf{\quad p↑ \quad}      & \textbf{MAE↓}    & \textbf{RMSE↓}   & \textbf{\quad p↑ \quad}     & \textbf{MAE↓}    & \textbf{RMSE↓}   & \textbf{\quad p↑ \quad}     & \textbf{MAE↓}    & \textbf{RMSE↓}   & \textbf{\quad p↑ \quad}     \\
\midrule 
UBFC & 
CHROM \cite{de2013robust}    & 0.2221 & 0.2817 & 0.0698 & 0.2221 & 0.2817 & 0.0698 & 0.6708  & 1.0542 & 0.1054 & 7.2291  & 8.9224  & 0.5123 \\
& POS \cite{wang2016algorithmic}     & 0.2364 & 0.2861 & 0.1359 & 0.2364 & 0.2861 & 0.1359 & 0.6515  & 0.9535 & 0.1345 & 7.3539  & 8.0402  & 0.4923 \\
& Rhythmnet \cite{niu2019rhythmnet} & 0.0621 & 0.0813 & 0.1873 & 0.0621 & 0.0813 & 0.1873 & 0.1985  & 0.2667 & 0.3043 & 5.1542  & 7.4672  & 0.8165 \\
& PulseGAN \cite{song2021pulsegan}     & 0.0638 & 0.0825 & 0.1883 & 0.0638 & 0.0825 & 0.1883 & 0.1961  & 0.2631 & 0.3018 & 5.2852  & 7.5116  & 0.8372 \\
& NEST \cite{lu2023neuron}  & 0.0597 & 0.0782 & 0.2017 & 0.0597 & 0.0782 & 0.2017 & 0.2138  & 0.2824 & 0.3179 & 4.7471  & 6.8876  & 0.8546 \\
& HSRD \cite{wang2023hierarchical}  & 0.0591 & 0.0748 & 0.2271 & 0.0591 & 0.0748 & 0.2271 & 0.2104  & 0.2715 & 0.3201& 4.5369  & 6.5612 & 0.8711 \\
& Bi-TTA \cite{li2024bi} & 0.0585 & 0.0721 & 0.2412 & 0.0585 & 0.0721 & 0.2412 & 0.1932 & 0.2749 & 0.3233 & 4.4159 & 6.3547 & 0.8730 \\
& \textbf{CiCi} & \textbf{0.0508}\(^*\) & \textbf{0.0717}\(^*\) & \textbf{0.2625}\(^*\) & \textbf{0.0508}\(^*\) & \textbf{0.0717}\(^*\) & \textbf{0.2625}\(^*\) & \textbf{0.1905}\(^*\) & \textbf{0.2714}\(^*\) & \textbf{0.3296}\(^*\) & \textbf{4.3905}\(^*\) & \textbf{6.1273}\(^*\) & \textbf{0.8863}\(^*\) \\
 \midrule 
PURE &CHROM \cite{de2013robust}   & 0.2096 & 0.2751 & 0.1059 & 0.2096 & 0.2751 & 0.0759 & 0.5404  & 0.8266 & 0.1173 & 9.7914  & 12.7568 & 0.3732 \\
& POS \cite{wang2016algorithmic}     & 0.1959 & 0.2571 & 0.1684 & 0.1959 & 0.2571 & 0.1684 & 0.5373  & 0.8460  & 0.1433 & 9.8273  & 13.4414 & 0.3432 \\
& Rhythmnet \cite{niu2019rhythmnet} & 0.0671 & 0.0923 & 0.6046 & 0.0671 & 0.0923 & 0.6046 & 0.2864  & 0.4184 & 0.5526 & 8.2542  & 11.1765 & 0.6832 \\
& PulseGAN \cite{song2021pulsegan} & 0.0635 & 0.0869 & 0.6107 & 0.0635 & 0.0869 & 0.6107  & 0.2853  & 0.4049 & 0.5575 & 8.2006  & 11.1563 & 0.6889 \\
& NEST \cite{lu2023neuron}  & 0.0635 & 0.0874 & 0.6422 & 0.0635 & 0.0874 & 0.6422 & 0.2255  & 0.3505 & 0.5734 & 7.6889 & 10.4783 & 0.7255 \\
& HSRD \cite{wang2023hierarchical}  & 0.0631 & 0.0852 & 0.6531 & 0.0631 & 0.0852 & 0.6531 & 0.2192  & 0.3414 & 0.5769 & 7.5618  & 10.2680 & 0.7425 \\
& Bi-TTA \cite{li2024bi} & 0.0617 & 0.0851 & 0.6513 & 0.0617 & 0.0851 & 0.6513 & 0.2173  & 0.3439 & 0.5836 & 7.5097 & 10.3249 & 0.7482 \\
& \textbf{CiCi} & \textbf{0.0606}\(^*\) & \textbf{0.0811}\(^*\) & \textbf{0.6690}\(^*\) & \textbf{0.0606}\(^*\) & \textbf{0.0811}\(^*\) & \textbf{0.6690}\(^*\) & \textbf{0.2048}\(^*\)  & \textbf{0.3565}\(^*\) & \textbf{0.6013}\(^*\) & \textbf{7.3771}\(^*\) & \textbf{10.032}\(^*\) & \textbf{0.7655}\(^*\) \\
  \midrule                     
BUAA & CHROM\cite{de2013robust}    & 0.3786 & 0.3237 & 0.0682 & 0.3786 & 0.3237 & 0.0682 & 0.6813  & 0.8836 & 0.0715 & 6.0934  & 8.2938  & 0.5165 \\
& POS \cite{wang2016algorithmic}      & 0.3198 & 0.3762 & 0.0962 & 0.3198 & 0.3762 & 0.0962 & 0.6275  & 0.8424 & 0.1127 & 5.0407  & 7.1198  & 0.6374 \\                      
& Rhythmnet \cite{niu2019rhythmnet} & 0.1451 & 0.1681 & 0.2891 & 0.1451 & 0.1681 & 0.2891 & 0.5564  & 0.6904 & 0.2914 & 3.7852  & 6.3237  & 0.7492 \\
&                       PulseGAN ~\cite{song2021pulsegan} & 0.1449 & 0.1681 & 0.2895 & 0.1449 & 0.1681 & 0.2895 & 0.5474  & 0.6901 & 0.2966 & 3.6411  & 6.1238  & 0.7502 \\
& NEST \cite{lu2023neuron}  & 0.1436 & 0.1665 & 0.2955 & 0.1436 & 0.1665 & 0.2955 & 0.5514  & 0.6884 & 0.3004 & 3.3723  & 5.8806  & 0.7647 \\
& HSRD \cite{wang2023hierarchical}  & 0.1395 & 0.1558 & 0.3014 & 0.1395 & 0.1558 & 0.3014 & 0.5207  & 0.6212 & 0.3109 & 2.8921  & 4.0723  & 0.7788 \\
& Bi-TTA \cite{li2024bi} & 0.1369 & 0.1424 & 0.3211 & 0.1369 & 0.1424 & 0.3211 & 0.5141 & 0.6107 & 0.3225 & 2.7362 & 4.0462 & 0.7891   \\
& \textbf{CiCi} & \textbf{0.1246}\(^*\) & \textbf{0.1371}\(^*\) & \textbf{0.3446}\(^*\)  & \textbf{0.1246}\(^*\) & \textbf{0.1371}\(^*\) & \textbf{0.3446}\(^*\) & \textbf{0.5038}\(^*\) & \textbf{0.6029}\(^*\) & \textbf{0.3375}\(^*\) & \textbf{2.6857}\(^*\) & \textbf{3.7598}\(^*\) & \textbf{0.7915}\(^*\)  \\
\bottomrule 
\end{tabular}
\label{table: HRV}
\end{table*}

\subsection{Visualization of BVP Signal in TTA}

\begin{figure}[ht] 
    \centering
    \includegraphics[width=0.48\textwidth]{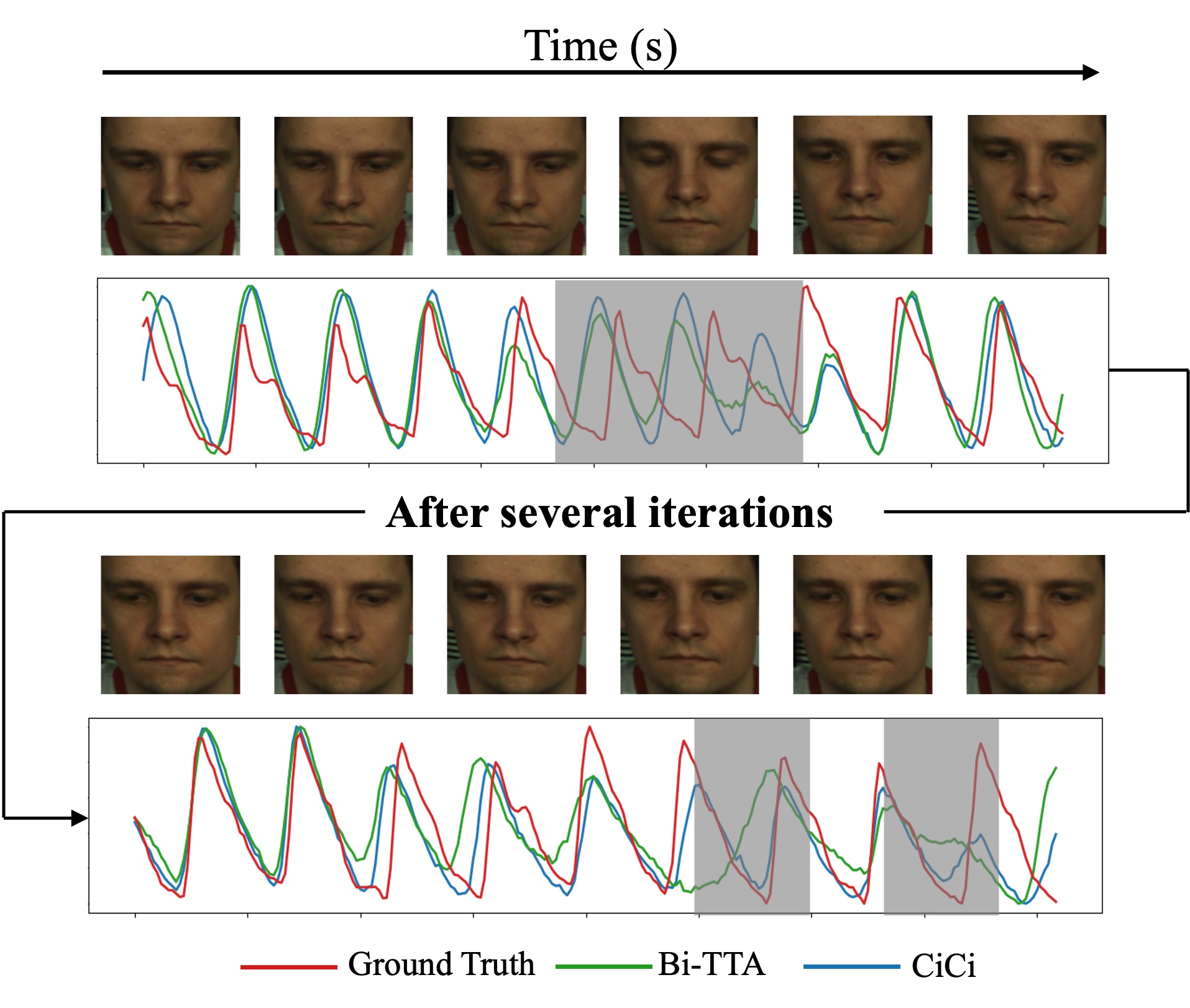} 
    \caption{The visualized performance of Bi-TTA and CiCi in TTA. We highlighted the regions where CiCi performed well with grey.}
    \label{fig:BVP}
\end{figure}

To more intuitively demonstrate the performance improvement of CiCi throughout the testing process, we visualized the BVP signals during the TTA procedure for the SOTA method, Bi-TTA, and our proposed CiCi method, as shown in Figure \ref{fig:BVP}. It can be observed that at the start of TTA, both Bi-TTA and our method produced outputs with blurred peaks. However, as TTA progressed, the performance of CiCi noticeably improved, highlighting the superiority of our approach.

\begin{table*}[ht]
\scriptsize
\setlength{\tabcolsep}{6pt}
\centering
\caption{HR estimation results with mmWave under TTA.}
\begin{tabular}{lcccccccccccc}
\toprule
  & \multicolumn{3}{c}{\textbf{Flat \& Unobstructed}} 
  & \multicolumn{3}{c}{\textbf{Flat \& Congested}} 
  & \multicolumn{3}{c}{\textbf{Bumpy \& Congested}}
  & \multicolumn{3}{c}{\textbf{EquiPleth}}\\
  \cmidrule(lr){2-4} \cmidrule(lr){5-7} \cmidrule(lr){8-10} \cmidrule(lr){11-13}
\textbf{Method} 
  & \textbf{MAE↓} & \textbf{RMSE↓} & \textbf{\quad p↑ \quad}    
  & \textbf{MAE↓} & \textbf{RMSE↓} & \textbf{\quad p↑ \quad}    
  & \textbf{MAE↓} & \textbf{RMSE↓} & \textbf{\quad p↑ \quad}
  & \textbf{MAE↓} & \textbf{RMSE↓} & \textbf{\quad p↑ \quad}\\
\midrule

RF-vital \cite{choi2022remote}
  & 5.94 & 7.71 & 0.69 
  & 6.07 & 7.84 & 0.67
  & 6.21 & 8.05 & 0.63 
  & 7.43 & 9.94 & 0.45\\

TENT \cite{wang2020tent} 
  & 5.69 & 7.50 & 0.70 
  & 5.87 & 7.63 & 0.69
  & 6.05 & 7.84 & 0.66 
  & 7.17 & 9.69 & 0.48\\

SAR \cite{niu2023towards}
  & 5.61 & 7.43 & 0.72 
  & 5.77 & 7.44 & 0.72
  & 5.96 & 7.63 & 0.69 
  & 7.03 & 9.58 & 0.51\\

TTT++ \cite{liu2021ttt++}
  & 5.54 & 7.26 & 0.75 
  & 5.71 & 7.32 & 0.73
  & 5.89 & 7.54 & 0.70 
  & 6.92 & 9.39 & 0.55\\

EATA \cite{niu2022efficient}
  & 5.49 & 7.14 & 0.76 
  & 5.68 & 7.22 & 0.74
  & 5.80 & 7.45 & 0.73 
  & 6.89 & 9.26 & 0.56\\
  
SHOT \cite{liang2021source}
  & 5.43 & 7.09 & 0.77 
  & 5.62 & 7.18 & 0.75
  & 5.69 & 7.33 & 0.75 
  & 6.68 & 9.15 & 0.58\\

DeYO \cite{leeentropy} 
  & 5.45 & 7.14 & 0.76 
  & 5.55 & 7.16 & 0.76
  & 5.61 & 7.28 & 0.74 
  & 6.72 & 9.23 & 0.57\\

Bi-TTA \cite{li2024bi}
  & 5.26 & 6.79 & 0.79 
  & 5.21 & 6.98 & 0.78
  & 5.43 & 7.18 & 0.75 
  & 6.54 & 9.16 & 0.60\\
\midrule 

\textbf{CiCi}  
  & \textbf{5.13} & \textbf{6.55}& \textbf{0.82} 
  & \textbf{5.18} &\textbf{6.67} & \textbf{0.81} 
  & \textbf{5.24} &\textbf{6.84} & \textbf{0.77} 
  & \textbf{6.42} &\textbf{8.83} & \textbf{0.64} \\
\bottomrule
\end{tabular}
\label{tabel: mmwave}
\end{table*}

\subsection{HRV Estimation}
Due to the absence of reliable BVP labels in VIPL and V4V, we conducted HRV experiments using the UBFC, PURE, and BUAA datasets, focusing on key metrics including LFnu, HFnu, and the LF/HF ratio. As shown in Tab. \ref{table: HRV}, we compared CiCi approach against traditional methods, deep learning techniques, and TTA methods, including the rPPG-specific Bi-TTA. Our findings show that TTA methods consistently outperformed other deep learning approaches and exceeded Bi-TTA performance across all metrics, indicating that STTI enhanced the model's ability to capture accurate pulse waveforms and adapt robustly to various scenarios.

\subsection{HR Estimation in mmWave}


Since CiCi requires no modifications to the model architecture, it can be directly applied to other RPM tasks. To evaluate CiCi’s performance in millimeter-wave (mmWave) physiological monitoring, we conducted experiments using data from PhysDrive \cite{wang2025physdrive}, combined with the EquiPleth dataset \cite{vilesov2022blending}. Specifically, PhysDrive provides mmWave data across three different driving scenarios of Flat \& Unobstructed, Flat \& Congested, and Bumpy \& Congested, corresponding flat and unobstructed roads, flat but congested roads, and rugged, congested roads, respectively. The varying levels of road bumpiness introduce different degrees of difficulty in fitting physiological signals from the mmWave data. We treated these as three separate datasets and trained them jointly with EquiPleth under the TTA protocol. The results are presented in Tab. \ref{tabel: mmwave}.

From the results, we can observe that the CiCi method demonstrates a significant performance advantage in the mmWave task. This suggests that, owing to its design based on physiological signal priors, CiCi holds strong potential for extension to a broader range of non-contact physiological signal monitoring applications beyond just RGB videos.

\section{Conclusion}

Overall, based on observations, we introduce a novel knowledge prior based on the inherent morphological inconsistencies in the time domain. The spatio-temporal inconsistencies in the time domain, coupled with the spatio-temporal consistencies in the frequency domain, jointly constrain the adapting process, surpassing the performance of all previous self-supervised learning methods in RPM. Additionally, by analyzing the gradients of both priors, we identified the reasons for model collapse during multiple iterations due to the inconsistency prior. The proposed gradient control strategy effectively enhances the model's stability and mitigates the model degradation phenomena when the conflict of loss function existed. The findings of our work can provide insights into future RPM method designs.

\bibliographystyle{IEEEtran}
\bibliography{IEEEabrv,reference}


 

\begin{IEEEbiography}[{\includegraphics[width=1in,height=1.25in,clip,keepaspectratio]{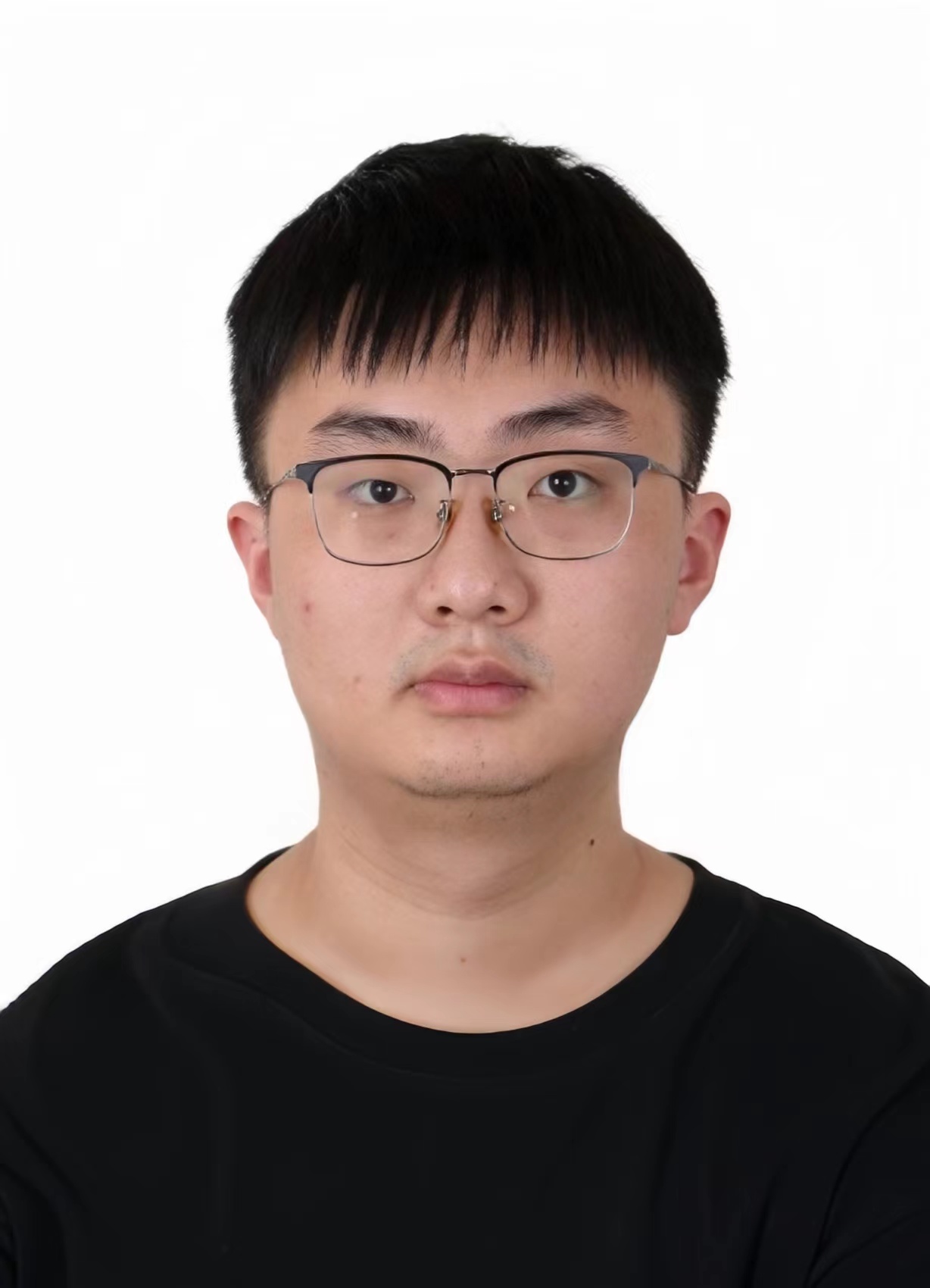}}]{Xiao Yang} is currently an M.Phil. student at the Hong Kong University of Science and Technology, Guangzhou campus. He received his bachelor's degree in Computing Science at Sichuan Agricultural University. His research interests include physiological signal measurement, state monitoring, and human factors.
\end{IEEEbiography}

\begin{IEEEbiography}[{\includegraphics[width=1in,height=1.25in,clip,keepaspectratio]{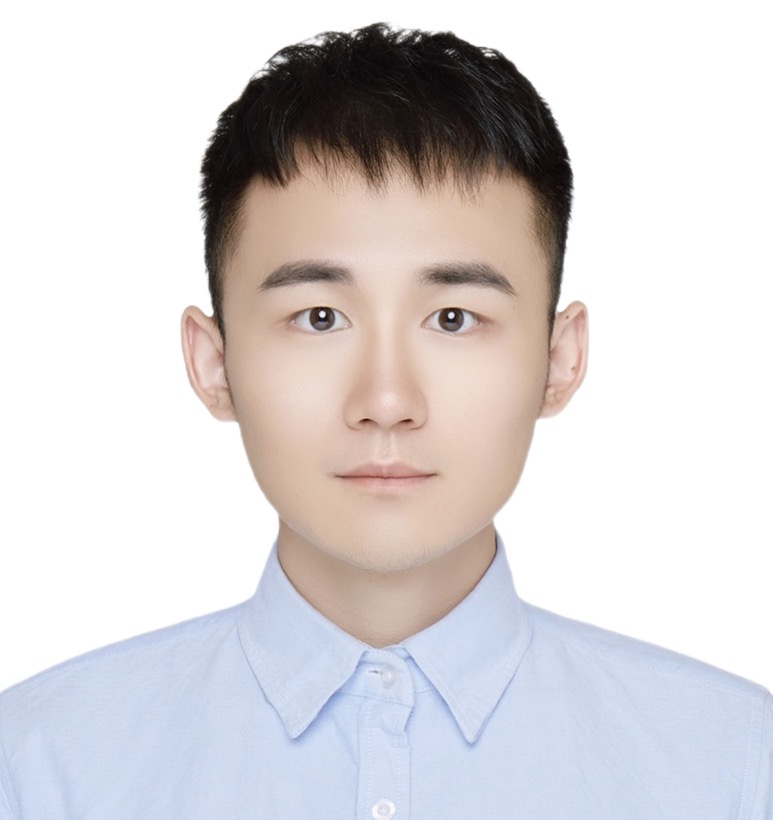}}]{Jiyao Wang} received the B.Eng. degree in Software Engineering from Sichuan University, Chengdu, China in 2021, M.Sc. degree in Big Data Technology from the Hong Kong University of Science and Technology (HKUST), Hong Kong S.A.R., China, in 2022, and a Ph.D. degree at HKUST, Guangzhou campus. Currently, he is a postdoctoral researcher at McGill University, Canada. His research interests include physiological signal measurement, intelligent transport systems, and human factors.
\end{IEEEbiography}

\begin{IEEEbiography}[{\includegraphics[width=1in,height=1.25in,clip,keepaspectratio]{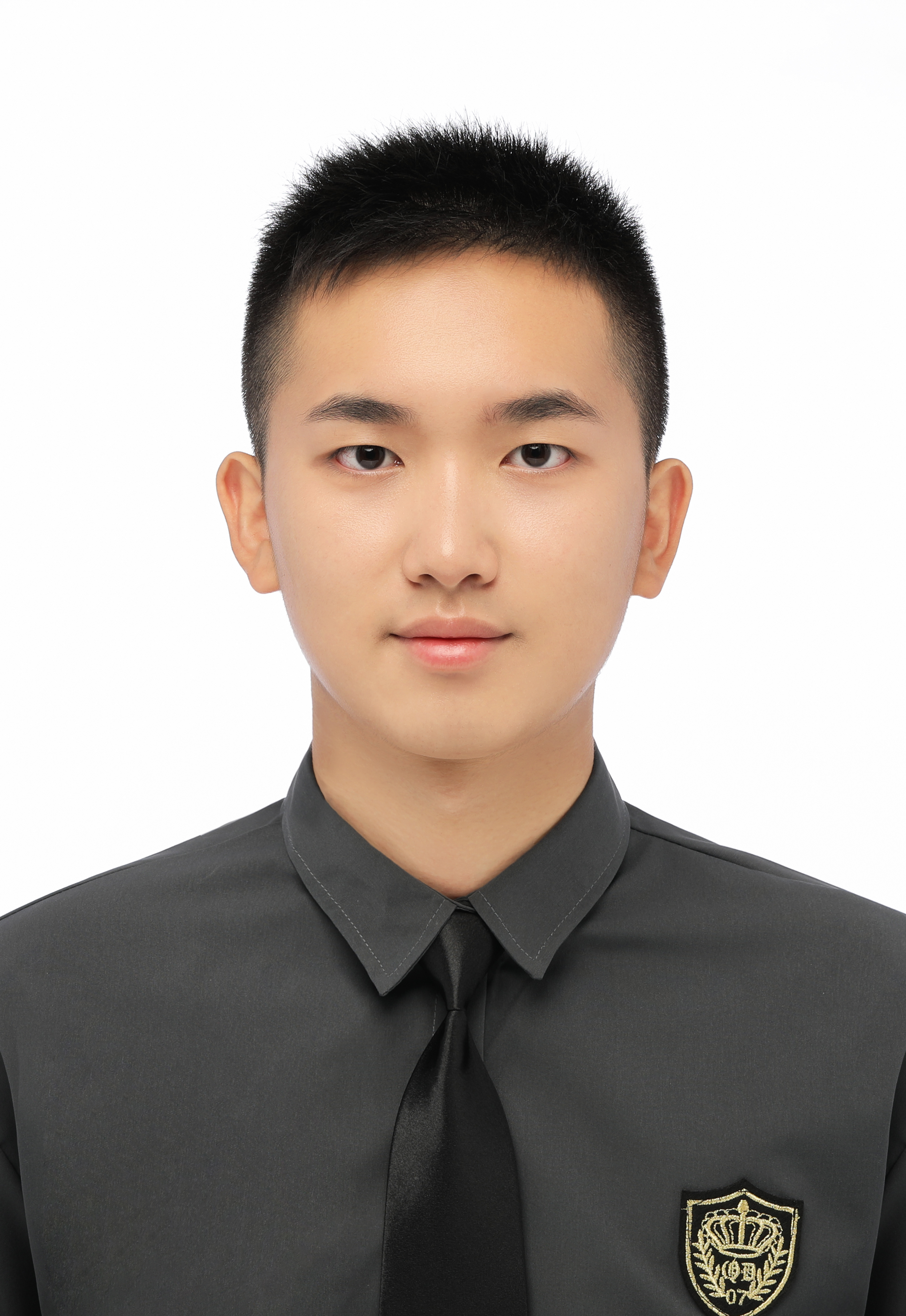}}]{Yuxuan Fan} obtained a bachelor's degree in School of Advanced Engineering at University of Science and Technology Beijing and is currently pursuing master degree at HKUST, Guangzhou campus. His research interests include MLLM and AI for Healthcare.
\end{IEEEbiography}

\begin{IEEEbiography}[{\includegraphics[width=1in,height=1.25in,clip,keepaspectratio]{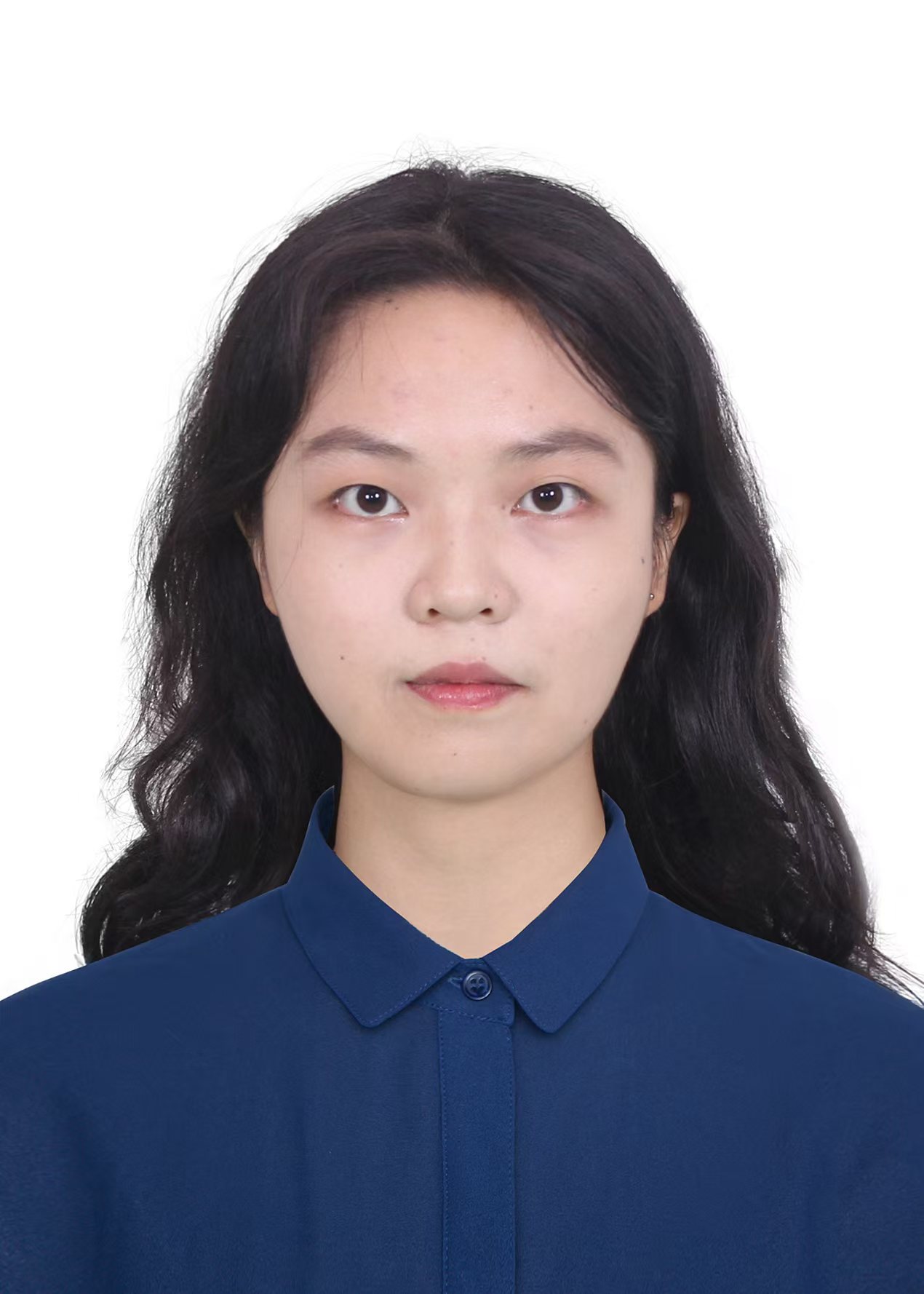}}]{Can Liu} is currently pursuing a bachelor's degree in Data Science and Big Data Technology at Sichuan Agricultural University. Her research interests include physiological measurement and multimodal analysis
\end{IEEEbiography}

\begin{IEEEbiography}[{\includegraphics[width=1in,height=1.25in,clip,keepaspectratio]{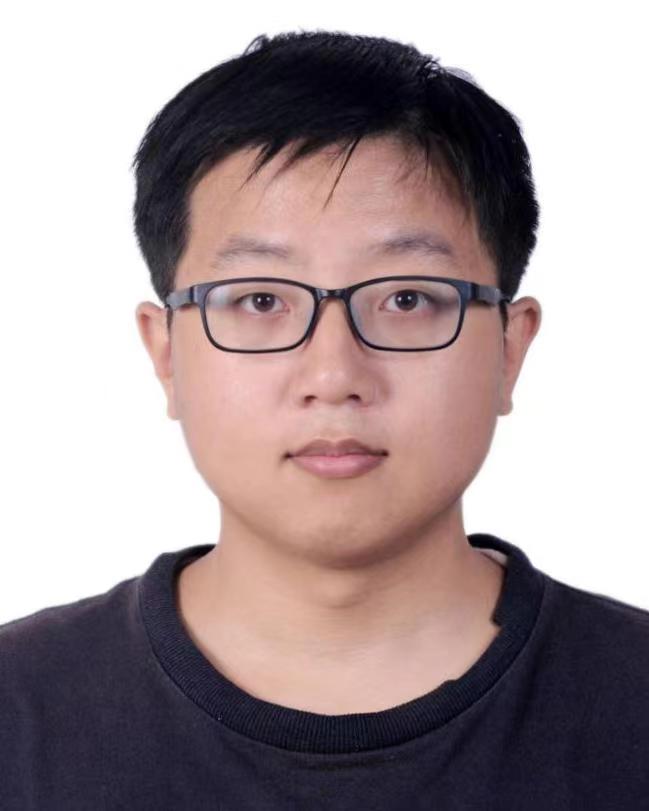}}]{Houcheng Su} is currently a Ph.D. student at the Hong Kong University of Science and Technology (Guangzhou). He received his M.S. degree from the University of Macau. His research interests include computer vision, transfer learning, medical imaging, and bioinformatics.
\end{IEEEbiography}

\begin{IEEEbiography}[{\includegraphics[width=1in,height=1.25in,clip,keepaspectratio]{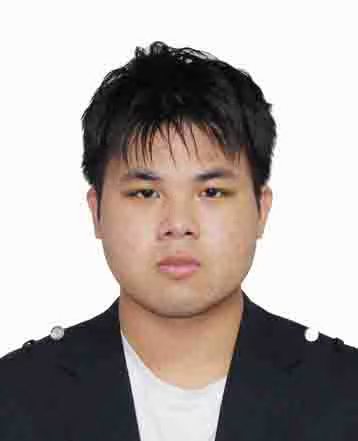}}]{Weichen Guo} is currently a bachelor student learning Data Science and Big Data at Sichuan Agricultural University. His main research directions include machine learning, semantic segmentation, and remote sensing.
\end{IEEEbiography}

\begin{IEEEbiography}[{\includegraphics[width=1in,height=1.25in,clip,keepaspectratio]{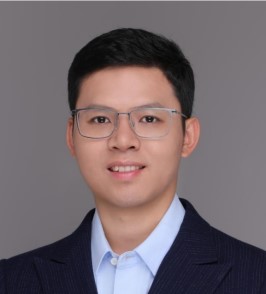}}]{Zitong Yu} (Senior Member, IEEE) received the Ph.D. degree in Computer Science and Engineering from the University of Oulu, Finland, in 2022. Currently, he is an Associate Professor at Great Bay University, China. He was a Postdoctoral researcher at ROSE Lab, Nanyang Technological University. He was a visiting scholar at TVG, University of Oxford. His research interests focus on subtle visual computing. He was area chairs of ACM MM 2025, ICME 2023, BMVC 2024, and IJCB 2024. He was a recipient of IAPR Best Student Paper Award, IEEE Finland Section Best Student Conference Paper Award 2020, Best Paper Candidate of ICME 2024, second prize of the IEEE Finland Jt. Chapter SP/CAS Best Paper Award (2022), and World's Top 2\% Scientists 2023/2024 by Stanford University.

\end{IEEEbiography}

\begin{IEEEbiography}[{\includegraphics[width=1in,height=1.25in,clip,keepaspectratio]{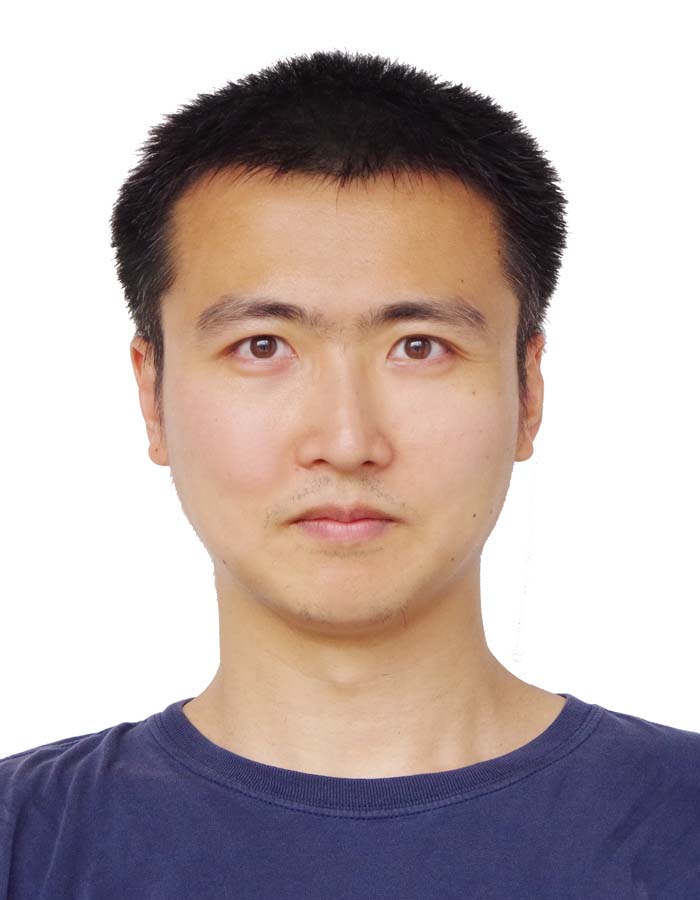}}]{Dengbo He} received his bachelor’s degree from Hunan University in 2012, M.S. degree from the Shanghai Jiao Tong University in 2016 and Ph.D. degree from the University of Toronto in 2020. He is currently an assistant professor from the Intelligent Transpiration Trust and Robotics and Autonomous Systems Thrust, the HKUST(Guangzhou). He is also affiliated with the Department of Civil and Environmental Engineering, HKUST, Hong Kong SAR. From 2020 to 2021, he was a post-doctoral fellow at the University of Toronto.
\end{IEEEbiography}

\begin{IEEEbiography}[{\includegraphics[width=1in,height=1.25in,clip,keepaspectratio]{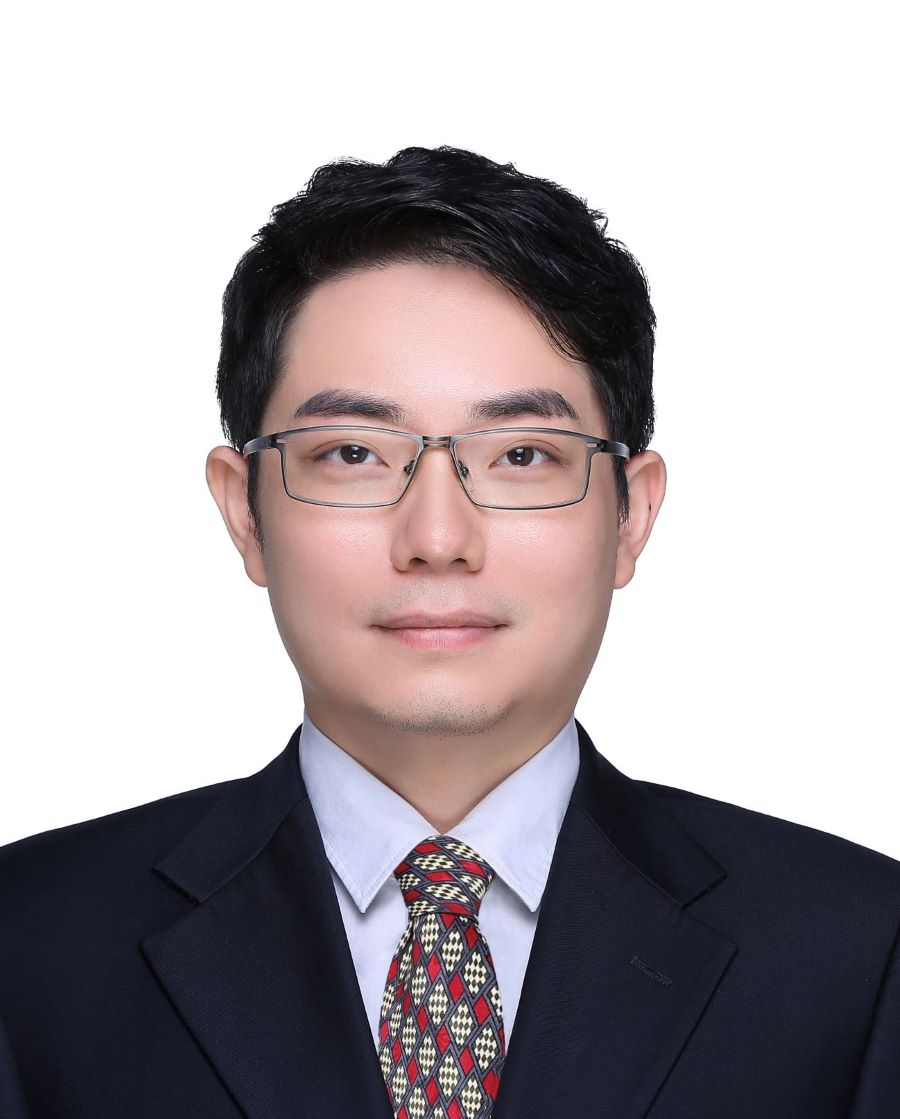}}]{Kaishun Wu}
(Fellow, IEEE) received the Ph.D. degree in computer science and engineering from HKUST, Hong Kong, in 2011. He was a Distinguished Professor and the Director of Guangdong Provincial Wireless Big Data and Future Network Engineering Center with Shenzhen University, Shenzhen, China. In 2022, he joined HKUST (GZ) as a Full Professor with DSA Thrust and IoT Thrust. He is an Active Researcher with more than 200 papers published on major international academic journals and conferences, as well as more than 100 invention patents, including 12 from the USA. He is an IET, AAIA, and IEEE Fellow.
\end{IEEEbiography}

\vspace{11pt}


\vfill

\end{document}